\def\year{2021}\relax
\documentclass[letterpaper]{article} 
\usepackage{aaai21}  
\usepackage{times}  
\usepackage{helvet} 
\usepackage{courier}  
\usepackage[hyphens]{url}  
\usepackage{graphicx} 
\usepackage{newtxmath}
\usepackage{natbib}  
\usepackage{caption} 
\usepackage[ruled,vlined]{algorithm2e}
\usepackage[noend]{algpseudocode}
\usepackage{adjustbox}
\usepackage{multirow}
\usepackage{siunitx}
\usepackage{textcomp}

\usepackage{svg}
\usepackage{caption}
\usepackage{subcaption}

\newcommand{\DSLR}{\texttt{\textbf{DSLR}}}
\newcommand{\DSLRUDA}{\texttt{\textbf{DSLR-UDA}}}
\newcommand{\DSLRSeg}{\texttt{\textbf{DSLR-Seg}}}
\newcommand{\DSLRpp}{\texttt{\textbf{DSLR++}}}

\newcommand{\Ach}{\texttt{\textbf{ADMG}}}
\newcommand{\CacciaAE}{\texttt{\textbf{CHCP-AE}}}   
\newcommand{\CacciaVAE}{\texttt{\textbf{CHCP-VAE}}} 
\newcommand{\CacciaGAN}{\texttt{\textbf{CHCP-GAN}}}
\newcommand{\Wu}{\texttt{\textbf{WCZC}}}    
\newcommand{\Bescos}{\texttt{\textbf{EmptyCities}}}

\title{DSLR: Dynamic to Static LiDAR Scan Reconstruction Using Adversarially Trained Autoencoder}
\author{
    Prashant Kumar\textsuperscript{\rm \thanks{denotes equal contribution}1},
    Sabyasachi Sahoo\textsuperscript{\rm \footnotemark[1]1},
    Vanshil Shah\textsuperscript{\rm 1} ,
    Vineetha Kondameedi\textsuperscript{\rm 1},
    Abhinav Jain\textsuperscript{\rm 1}, 
    Akshaj Verma\textsuperscript{\rm 1}, 
    Chiranjib Bhattacharyya\textsuperscript{\rm 1},
    \textnormal{and} Vinay Viswanathan \textsuperscript{\rm 2}\textsuperscript{\rm 3} \leavevmode \\
}

\affiliations{
    \textsuperscript{\rm 1}Indian Institute of Science, Bangalore, India \\
    \textsuperscript{\rm 2}AMIDC Pvt Ltd, Bangalore, India\\
    \textsuperscript{\rm 3}Chennai Mathematical Institute, Chennai, India\\
    \{prshnttkmr, ssahoo.iisc, vanshilshah, vineetha.knd92, abhinav98jain, akshajverma7\}@gmail.com, chiru@iisc.ac.in, vinay@atimotors.com

}

\begin{document}

\maketitle

\begin{abstract}
Accurate reconstruction of static environments from LiDAR scans of scenes containing dynamic objects, which we refer to as Dynamic to Static Translation (DST), is an important area of research in Autonomous Navigation. This problem has been recently explored for \textit{visual SLAM}, but to the best of our knowledge no work has been attempted to address DST for LiDAR scans. The problem is of critical importance due to wide-spread adoption of LiDAR in Autonomous Vehicles.  We show that state-of the art methods developed for the visual domain when adapted for LiDAR scans perform poorly.

We develop \DSLR{}, a deep generative model which learns a mapping between dynamic scan to its static counterpart through an adversarially trained autoencoder. Our model yields the first solution for DST on LiDAR that generates static scans without using explicit segmentation labels. 
\DSLR{} cannot always be applied to real world data due to lack of paired dynamic-static scans. Using Unsupervised Domain Adaptation, we propose \DSLRUDA{} for  transfer to real world data and experimentally show that this performs well in real world settings. Additionally, if segmentation information is available, we extend \DSLR{} to \DSLRSeg{} to further improve the reconstruction quality. 

\DSLR{} gives the state of the art performance on simulated and real-world datasets and also shows at least $4\times$ improvement. We show that \DSLR{}, unlike the existing baselines, is a practically viable model with its reconstruction quality within the tolerable limits for tasks pertaining to autonomous navigation like SLAM in dynamic environments.

\end{abstract}

\section{Introduction}
\noindent The problem of dynamic points occluding static structures is ubiquitous for any visual system. Throughout the paper, we define points falling on movable objects (e.g. cars on road) as dynamic points and the rest are called static points  \cite{chen2019suma++,ruchti2018mapping}. Ideally we would like to replace the dynamic counterparts by corresponding static ones. We call this as Dynamic to Static Translation problem. Recent works attempt to solve DST for the following modalities: Images \cite{bescos2019empty}, RGBD \cite{bevsic2020dynamic}, and point-clouds \cite{wu2020multimodal}. DST for LiDAR has also been attempted using geometric (non-learning-based) methods \cite{gskim-2020-iros,biasutti2017disocclusion}. To the best of our knowledge, no learning-based method has been proposed to solve DST for LiDAR scans.
\begin{figure}[t]
    \centering
    \includegraphics[scale=0.15]{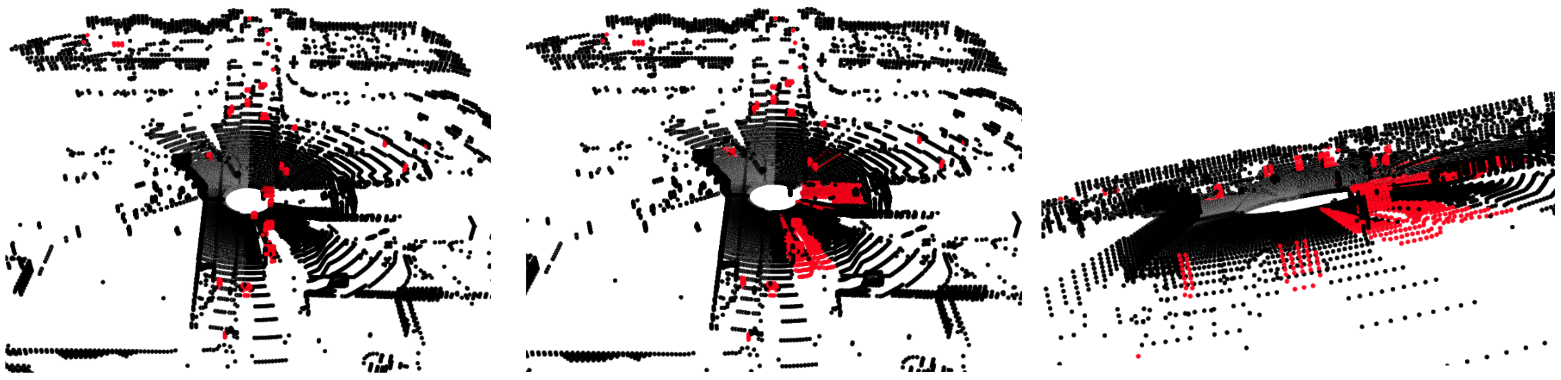}
    \caption{Static background points are shown in black. 1) Left: Dynamic LiDAR scan with occluding dynamic points in red. 2) Center: Static LiDAR scan reconstructed by DSLR. Although DSLR reconstructs complete scan only in-painted dynamic points are shown in red. 3) Right: Reconstructed static LiDAR scan shown from a different angle to highlight reconstruction quality of 3D structures like walls.}
    \label{slamtraject1}
\end{figure}
We show that existing techniques for non-LiDAR data produce sub-optimal reconstructions for LiDAR data \cite{caccia2018deep,achlioptas2017representation,groueix2018papier}. However, downstream tasks such as SLAM in a dynamic environment require accurate and high-quality LIDAR scans. 

To address these shortcomings, we propose a Dynamic to Static LiDAR Reconstruction (\DSLR{}) algorithm. It uses an autoencoder that is adversarially trained using a discriminator to reconstruct corresponding static frames from dynamic frames. Unlike existing DST based methods, \DSLR{} doesn't require specific segmentation annotation to identify dynamic points. We train \DSLR{} using a dataset that consists of corresponding dynamic and static scans of the scene. However, these pairs are hard to obtain in a simulated environment and even harder to obtain in the real world. To get around this, we also propose a new algorithm to generate appropriate dynamic and static pairs which makes \DSLR{} the first DST based model to train on a real-world dataset.\\

We now summarize the contributions of our paper:

\begin{itemize}
 \setlength{\itemsep}{0pt}
    \item  This paper initiates a study of DST for LiDAR scans and establishes that existing methods, when adapted to this 
    problem, yield poor reconstruction of the underlying static environments. This leads to extremely poor performance in  down-stream applications such as SLAM. To address this research gap we develop \DSLR{}, an adversarially trained autoencoder which learns a mapping from the latent space of dynamic scans to their static counterparts. This mapping is made possible due to the use of a \textit{pair-based discriminator}, which is able to distinguish between corresponding static and dynamic scans.  Experiments on simulated and real-world datasets show that \DSLR{} gives at least a $4\times$ improvement over adapted baselines. 
    
    \item \DSLR{} does not require segmentation information. However, if segmentation information is available, we propose an additional variant of our model \DSLRSeg{}. This model leverages the segmentation information and achieve an even higher quality reconstruction. To ensure that \DSLR{} works in scenarios where corresponding dynamic-static scan pairs might not be easily available, we utilise methods from unsupervised domain adaptation to develop \DSLRUDA{}. 
    
    \item  We show that \DSLR{}, when compared to other baselines, is the only model to have its reconstruction quality fall within the acceptable limits for SLAM performance as shown in our experiments. 

    \item  We open-source 2 new datasets, \textbf{CARLA-64}, \textbf{ARD-16} (Ati Realworld Dataset) consisting of corresponding static-dynamic LiDAR scan pairs for simulated and real world scenes respectively. We also release our pipeline to create such datasets from raw static and dynamic runs. \footnote{Code, Dataset and Appendix: https://dslrproject.github.io/dslr/}
\end{itemize}

\section{Related Work}
\label{related}

Point Set Generation Network \cite{fan2017point} works on 3D reconstruction using a single image, resulting in output point set coordinates that handle ambiguity in image ground truth effectively. PointNet \cite{qi2017pointnet} enables direct consumption of point clouds without voxelization or other transformations and provides a unified, efficient and effective architecture for a variety of down stream tasks. Deep generative models \cite{achlioptas2017representation} have attempted 3D reconstruction for point clouds and the learned representations in these models outperform existing methods on 3D recognition tasks. AtlasNet \cite{groueix2018papier} introduces an interesting method to generate shapes for 3D structures at various resolutions with improved precision and generalization capabilities. 

Recent approaches also explore the problem of unpaired point cloud completion using generative models \cite{chen2019unpaired}. They demonstrate the efficacy of their method on real life scans like ShapeNet. They further extend their work to learn one-to-many mapping between an incomplete and a complete scan using generative modelling \cite{wu2020multimodal}. Deep generative modelling for LiDAR scan reconstruction was introduced through the use of Conditional-GAN \cite{caccia2018deep}. The point clouds generated in LiDAR scans are different from normal point clouds. Unlike the latter, they give a 360\textdegree  view of the surrounding and are much more memory intensive, rich in detail and complex to work with. 
While most deep generative models for 3D data focus on 3D reconstruction, few attempts have been made to use them for DST. Empty-Cities \cite{bescos2018dynaslam} uses a conditional GAN to convert images with dynamic content into realistic static images by in-painting the occluded part of a scene with a realistic background.

\section{Methodology}
 Given a set of dynamic frames $D = \{d_{i}: i=1,\dots,n\}$, and their corresponding static frames $S = \{s_{i}:i=1,\dots,n \}$, our aim is to find a mapping between the latent space of dynamic frames to its corresponding static frames while preserving the static-structures in  $D_{i}$ and in-painting the regions occluded by dynamic-objects with the static-background. Our model consists of 3 parts: (a) An autoencoder trained to reconstruct LiDAR point clouds \cite{caccia2018deep}, (b) A pair discriminator to  distinguish ($S_{i}, S_{j}$) pairs from ($S_{i}, D_{j}$). Inspired by \cite{denton2017unsupervised}, the discriminator is trained using latent embedding vector pairs obtained using standard autoencoder as input, (c) An adversarial model that uses the above 2 modules to learn an autoencoder that maps dynamic scans to corresponding static scans. 

 We also describe $2$ variants of our models for (a) unsupervised domain adaptation to new environments, (b) utilizing segmentation information if available. For all following sections, x refers to a LiDAR range image (transformed to a  single column vector), $r(x)$ represent the the latent representation, $\overline{x}$ represents the reconstructed output,
 x\textsuperscript{i} refers to a component of x.

\subsection{DSLR: Dynamic to Static LiDAR Scan Reconstruction}

\begin{figure}[t]
    \centering
    \includegraphics[scale=0.25]{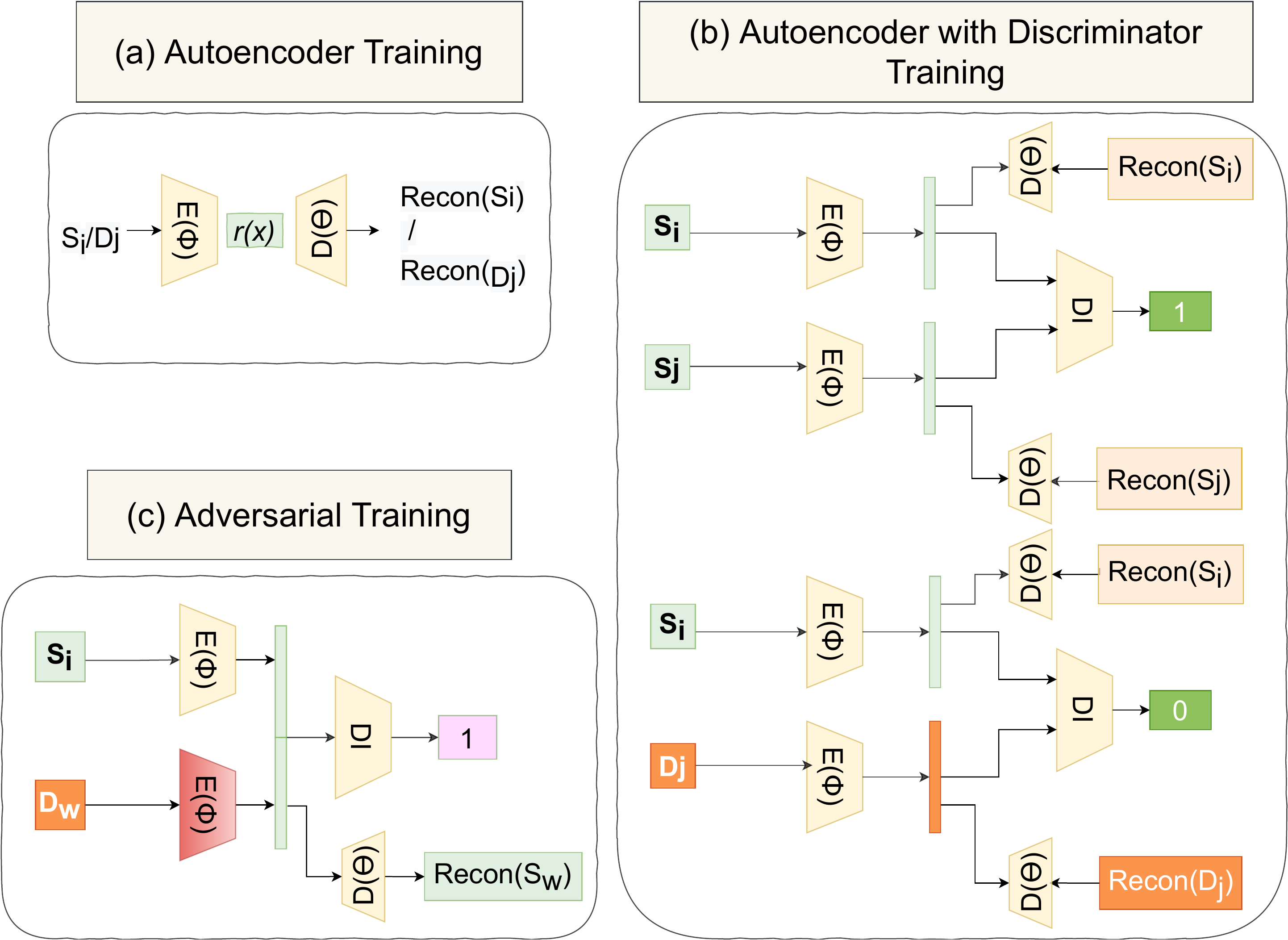}
    \caption{Model architecture. (a) Autoencoder (G) is trained to reconstruct LiDAR point clouds. (b) Discriminator (DI) is trained to discriminate $(x \in S, y \in S), (x \in S, y \in D)$ LiDAR scan pairs (c) Adversarial training using G and DI where $D_{w}$ and $S_{i}$ are corresponding dynamic-static pairs.}
    \label{fig:modelimp}
\end{figure}

\begin{figure}[t]
    \centering
    \vspace{5mm}
    \includegraphics[scale=0.28]{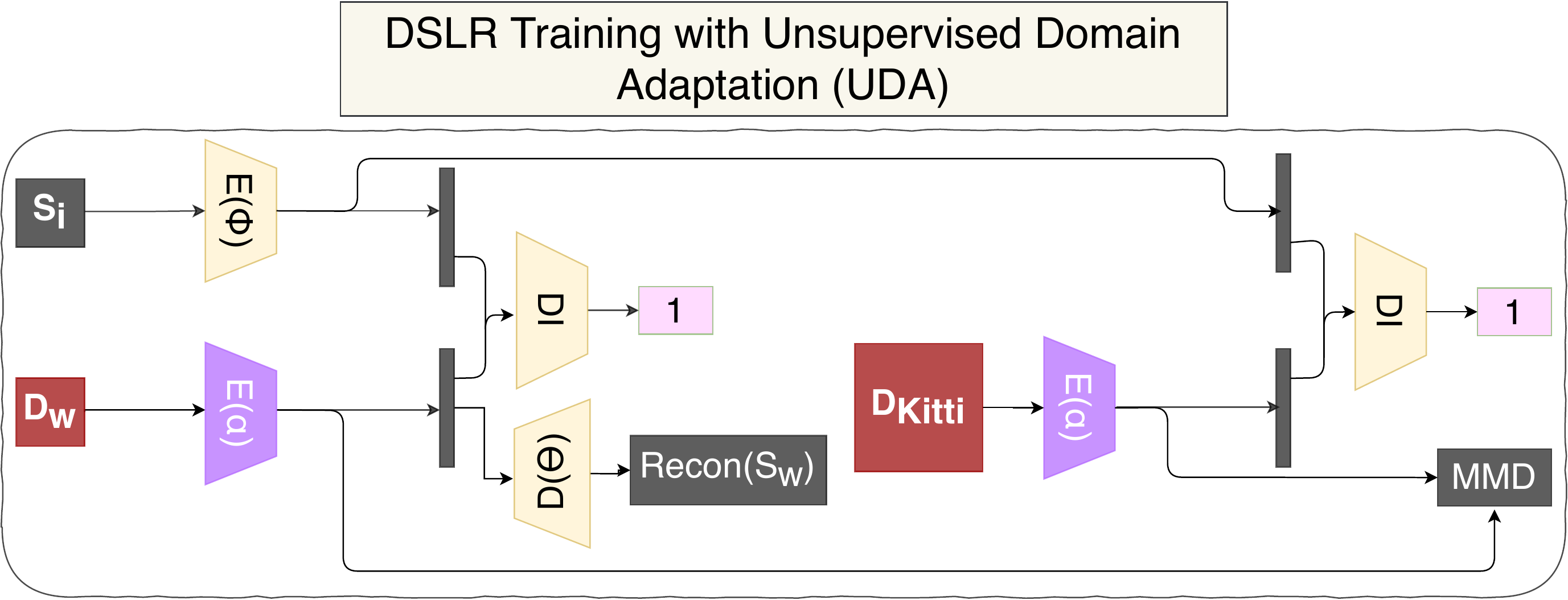}
    \caption{Adversarial training along with UDA for domain adaptation on KITTI data. Purple color indicate trainable weights. Refer Section \ref{uda}.}
    \label{fig:udamodel}
\end{figure}

The details of our model architecture and pipeline are illustrated in Fig. \ref{fig:modelimp}.

\subsubsection{Input Preprocessing:}
We use the strategy described by \cite{caccia2018deep} for converting LiDAR point clouds into range images. 

\subsubsection{Autoencoder:}
The Encoder ($E_{\phi}$) and Decoder ($D_{\theta}$) are based on the discriminator and generator from DCGAN \cite{radford2015unsupervised} as adapted in \cite{caccia2018deep}. However, we change the model architecture to a \(40 \times 512 \) grid instead of a \( 64 \times 1024 \) grid. This is done to discard the outer circles in a LiDAR scan because it contains the most noise and have least information about the scene. The bottleneck dimension of our model is $160.$ We define the network G as,
     \begin{equation}
     G : x \xrightarrow[]{E_{\phi}} r(x) \xrightarrow[]{D_{\theta}} \overline{x}
     \quad \end{equation}
    
The autoencoder $G_{\phi,\theta}$ is trained with  all s\textsubscript{i}, d\textsubscript{j} to reconstruct a given input at the decoder. We use a pixel-wise reconstruction loss for the LiDAR based range-image.
\begin{equation}
    \operatorname{MSE}(x,\overline{x})=\| x- \overline{x}\|^2 
\end{equation}

The autoencoder tries to learns the optimal latent vector representations $r(x)$ for static and dynamic LiDAR scans for efficient reconstruction. For ease of representation in the rest of the paper we omit the usage of parameters $\phi$ and $\theta$ for encoder and decoder of $G$, and they should be considered implicitly unless specified otherwise. 
    
\subsubsection{Adversarial Model}
    \label{Disc}
    \setlength{\itemsep}{0pt}
    \subsubsection{Pair Discriminator}
\label{PairDisc}

We use a feed-forward network based discriminator (DI). DI takes random scan pairs ($s_{i}, s_{j}$) and ($s_{i}$, $d_{j}$) and transforms them to latent vector pairs using $G.$ It is trained to output $1$ if both latent vectors represent static LiDAR scans, else $0$.

\begin{equation}
DI\left(r(x_{1}), \left(r(x_{2}\right)\right)=\left\{\begin{array}{ll}
1 &   x_{1} \in S, x_{2} \in S  \\
0 &   x_{1} \in S, x_{2} \in D \\
\end{array}\right\}
\label{discequation}
\end{equation}

The discriminator is trained using Binary Cross-Entropy (BCE) Loss between $\overline{y}$ and $y$, where  $\overline{y}$ denotes the discriminator output and $y$ denotes the ground truth.

Corresponding static and dynamic LiDAR frames look exactly the same except for the dynamic-objects and holes present in the latter.  It is hard for the discriminator to distinguish between these two because of minimal discriminatory features present in the pairs. To overcome this, we use a dual-loss setting where we jointly train the already pre-trained autoencoder G along with discriminator. Therefore, the total loss is the sum of MSE Loss of autoencoder and BCE loss of discriminator.  Using reconstruction loss not only helps in achieving training stability but also forces the encoder to output a latent vector that captures both generative and discriminative features of a LiDAR scan. We represent the loss as $L\textsubscript{DI}.$ For simplicity, we present $L\textsubscript{DI}$ for only one set of input data ($x_{1}, {x}_{2}, {x}_{3}$), where ${x}_{1}, {x}_{2} \in S$, ${x}_{3} \in D$.

\begin{equation}
\begin{split}
    L_{DI} = & MSE({x}_{1},\overline{{x}_{1}}) + MSE({x}_{2},\overline{{x}_{2}}) +  MSE({x}_{3},\overline{{x}_{3}})\\ & + BCE(DI(r({x}_{1}) , r({x}_{2})), 1) \\ &
    + BCE(DI(r({x}_{1}), r({x}_{3})), 0)
\end{split}
\label{DILoss}
\end{equation}

\subsubsection{Adversarial Training:}
In the adversarial training, we create two copies of the autoencoder represented as $G^1_{\phi_{1},\theta_{1}}$ and $G^2_{\phi_{2},\theta_{2}}$. 
We freeze $\phi_1$, $\theta_1$, $\theta_2$, leaving only $\phi_{2}$ of $G^2$ trainable. 
Inputs ${{x}_{1} \in S, {{x}_{2}} \in D}$ are passed through $G^1$ and $G^2$ respectively, which output $r({x}_{1})$ and $r({x}_{2})$.
These outputs are the latent representations of the static and dynamic scans. We concatenate $r({x}_{1})$ and $r({x}_{2})$ and feed it to the discriminator. DI should ideally output $0$ as shown in Eq.~\ref{discequation}. However, in an adversarial setting, we want to fool the discriminator into thinking that both the vectors belong to $S$ and hence backpropagate the BCE loss w.r.t target value $1$, instead of $0.$ As training progresses the encoder weights for $G^2$, i.e. $\phi_2$, are updated to produce a static latent vector for a given dynamic frame, learning a latent vector mapping from dynamic scan to static scan. Additionally, we also backpropagate the reconstruction loss MSE(${x}_{2}$, $\overline{{x}_{2}}$), which ensures that $G^2$ generates the nearest corresponding static latent representation $z$ when given dynamic frame ${x}_{2}$ as input. The corresponding static scan reconstruction obtained from $z$ using $\theta_2$ in such a setting is qualitatively and quantitatively better than what would have been obtained through a simple dynamic-to-static reconstruction based autoencoder, as shown in Table \ref{table:mainresults}. We represent the adversarial loss as L\textsubscript{A}. Here, M denotes the total number of latent vector pairs used to train DI.
\begin{equation}
\begin{split}
    L_{A}(s_{i},d_{j})=\sum_{i=1}^{M}\sum_{j=i+1}^{M}-\log(DI(r(s_{i}), r(d_{j}))) \\ + MSE(s_{i},\overline{d_{j}})
\end{split}
\end{equation}

\begin{figure}[t]
    \centering
    \includegraphics[scale=0.32]{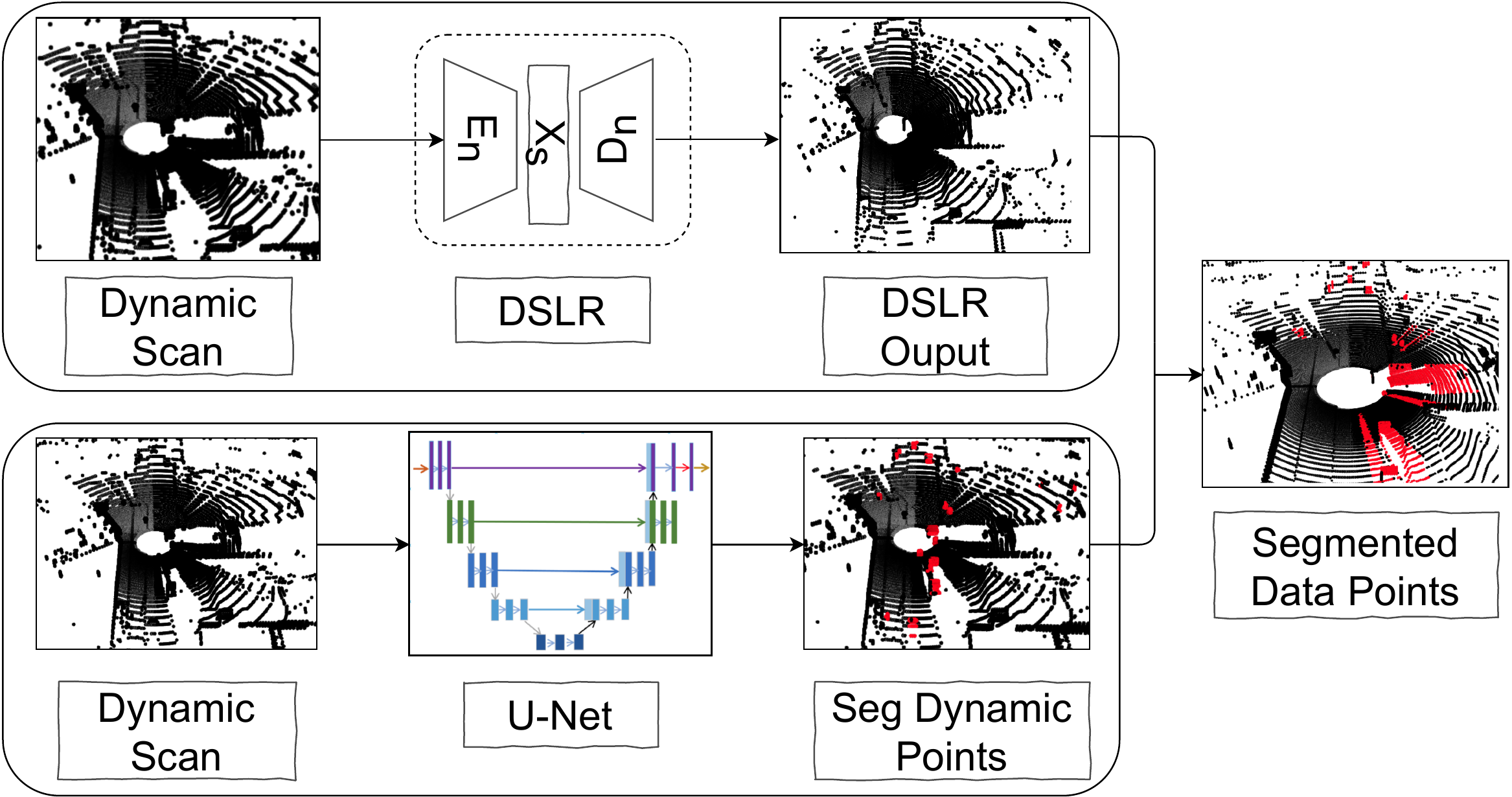}
    \caption{DSLR-Seg For a given dynamic frame, DSLR gives us static reconstruction and we use U-Net to get a static segmentation mask. We substitute dynamic points with reconstructed static points and take static points from the given dynamic frame for the final reconstruction.}
    \label{segmentmodel}
\end{figure}

\subsection{DSLR-UDA} \label{uda}

Due to the unavailability of ground truth static scans S for corresponding D, training \DSLR{} on real-world data is often not possible. A model trained on a simulated dataset usually performs poorly on real-world datasets due to the inherent noise and domain shift. 

To overcome this, we use Unsupervised Domain Adaptation (UDA) for adapting \DSLR{} to real world datasets where paired training data is unavailable.

In UDA, we have two domains: source and target. Source domain contains the output labels while the target domain does not. We take inspiration from \cite{tzeng2014deep} that uses a shared network with a Maximum Mean Discrepancy (MMD) \cite{Borgwardt2006IntegratingSB} loss that minimizes the distance between the source and target domain in the latent space. 

MMD loss is added to the adversarial phase of the training as shown in Fig. \ref{fig:udamodel}. Latent vectors from the dynamic-source and dynamic-target scans are used to calculate the MMD loss. For more details refer to Section 1.5 in Appendix\footnotemark[1]. Latent representations of static-source and target-dynamic scans are also fed to the discriminator with adversarial output as $1$, instead of $0$. All weights except for the encoder in blue, in Fig. \ref{fig:udamodel}, are frozen. The following equation denotes adversarial loss with UDA, L\textsubscript{U}:

\begin{equation}
\begin{split}
L_{U} = & L_{A}({s}_i,{d}_j) + \lambda \sum_{i=1}^{M} \sum_{j=i+1}^{M} MMD^{2}(r({d}_{j}), r({k}_{j}))
\end{split}
\end{equation}

where, $\textbf{s}_{i}$ $\in$ $S$,  ${d}_{j}$ $\in$ $D$ in source domain, and ${k}_{j}$ represents dynamic scan in target domain. $\lambda$ is a constant which is set to 0.01.

\subsection{DSLR-Seg}
Although \DSLR{} achieves a high-quality of reconstruction without segmentation information, it can leverage the same to further improve the reconstruction quality. High-quality LiDAR reconstructions are desirable in many downstream applications (e.g SLAM) for accurate perception of the environment. While the output reconstructions of \DSLR{} replace dynamic objects with corresponding static background accurately, it adds some noise in the reconstructed static which is undesirable.

To overcome this, we propose \DSLRSeg{}, which uses point-wise segmentation information to fill the dynamic occlusion points with the corresponding background static-points obtained from \DSLR{}. To this end, we train a U-Net \cite{ronneberger2015u} to segment dynamic and static points in a given LiDAR frame. The U-Net model outputs a segmentation mask.

We consider static-points from the input dynamic-frame and dynamic-points from the reconstructed static-output and generate the reconstructed output as shown in the Figure \ref{segmentmodel}. Mathematically, we can represent the above using Eq. \ref{eq-DSLR-seg}. 

\begin{equation}
    Recon  =  Mask_i*\overline{s_i} + (1-Mask_{i})d_i
\label{eq-DSLR-seg}
\end{equation}
Here, $Mask_i$ is a segmentation mask generated by U-Net model consisting values 0 for static and 1 for dynamic points.  $d_i$ is the input dynamic frame to the model and  $\overline{s_i}$ is the reconstructed static frame which is given by the model.

\begin{figure}[t]
    \centering
    \includegraphics[scale=0.20]{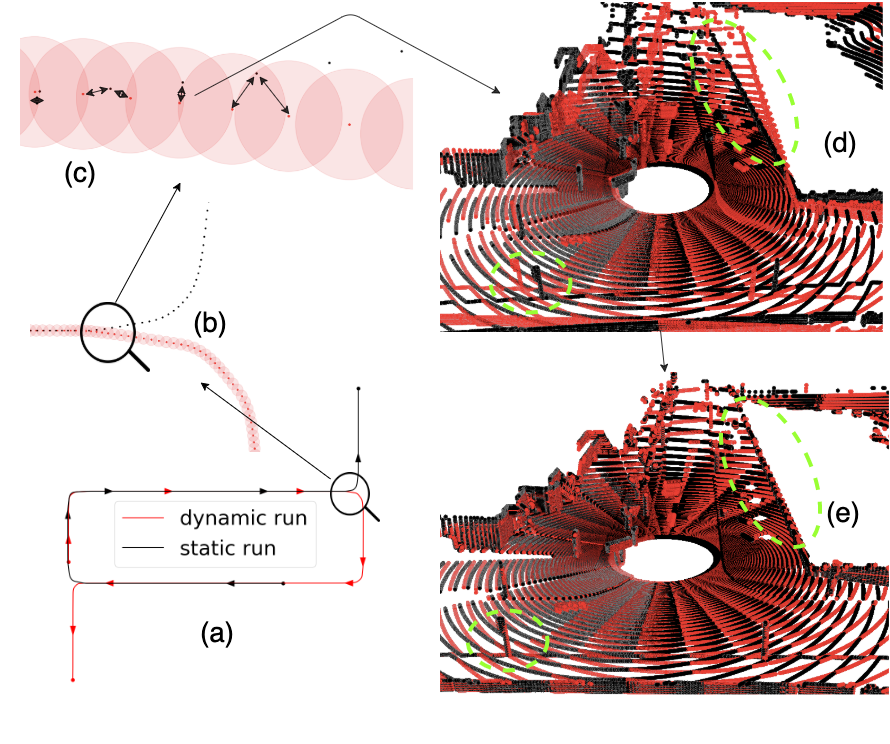}
    
    \caption{Dataset generation framework. Dynamic scans are shown in red, static scans in black. (a) shows a random dynamic and static run with overlap for few LiDAR poses. (b) zoomed-in portion of an overlapping region. (c) zoomed-in region showing paired correspondences using a bidirectional arrow. (d) shows a dynamic-static paired correspondence which often has significant mismatch (highlighted). (e) scan pair after applying pose transformation.}
    \label{dataset_building}
\end{figure}

\begin{figure*}[t]

\begin{subfigure}[t]{0.12\textwidth}
  \includegraphics[trim=100 0 0 0,clip,width=\linewidth]{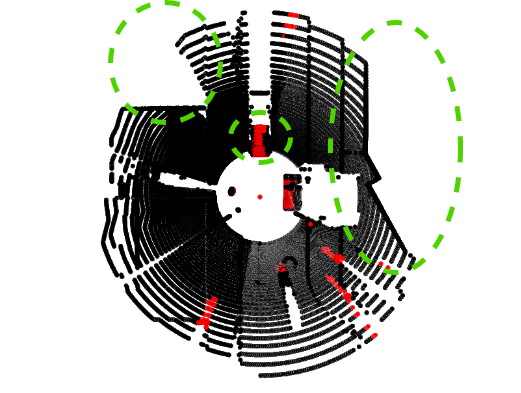}
  \label{fig:1}
\end{subfigure}\hfil 
\begin{subfigure}[t]{0.14\textwidth}
  \includegraphics[trim=180 100 100 100,clip,width=\linewidth]{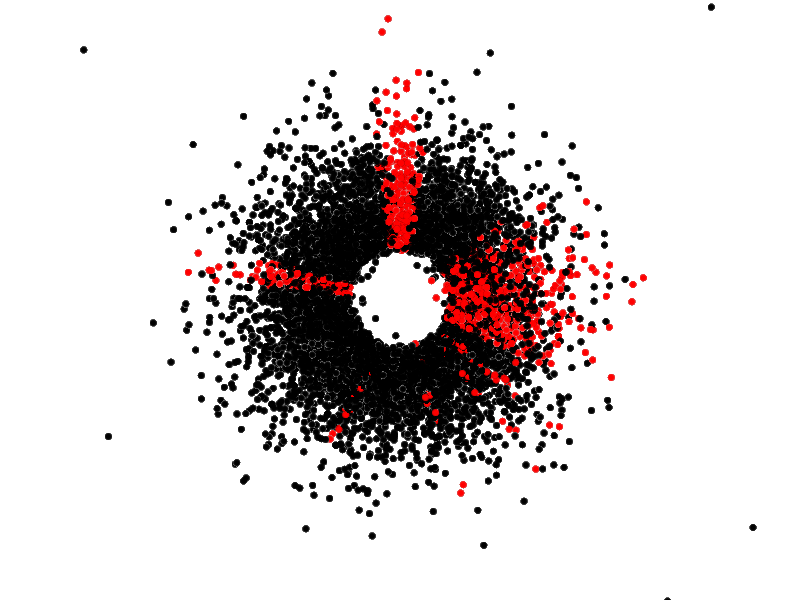}
  \label{fig:19}
\end{subfigure}\hfil
\begin{subfigure}[t]{0.129\textwidth}
  \includegraphics[trim=150 20 20 20,clip,width=\linewidth]{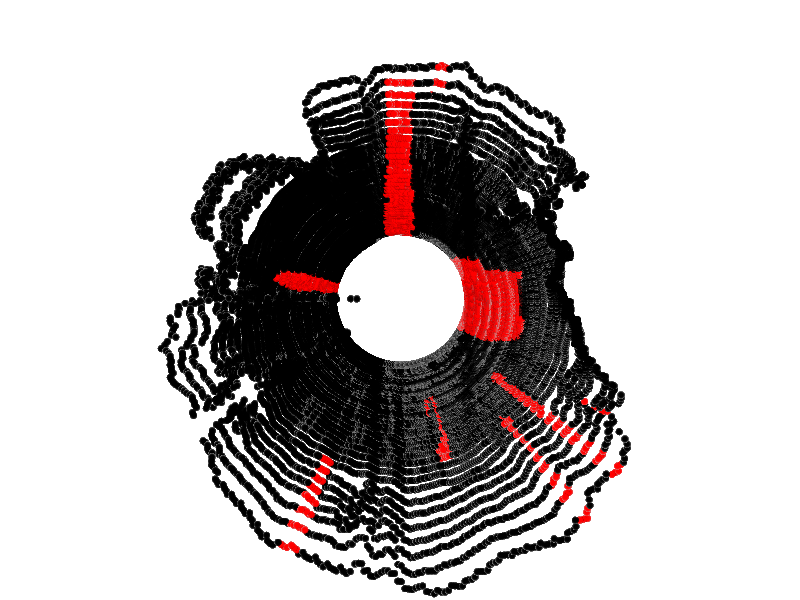}
  \label{fig:4}
\end{subfigure}\hfil
\begin{subfigure}[t]{0.115\textwidth}
  \includegraphics[trim=180 20 60 20,clip,width=\linewidth]{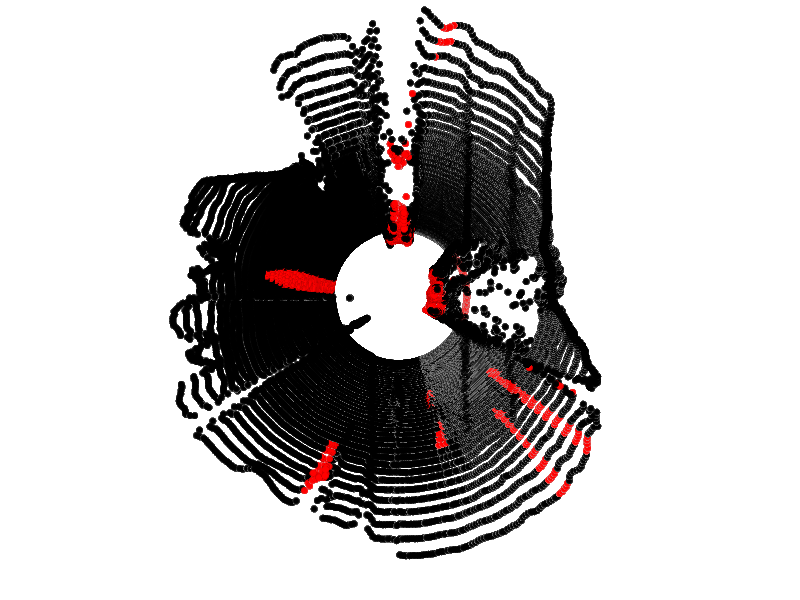}
  \label{fig:5}
\end{subfigure}\hfil
\begin{subfigure}[t]{0.115\textwidth}
  \includegraphics[trim=220 20 100 20,clip,width=\linewidth]{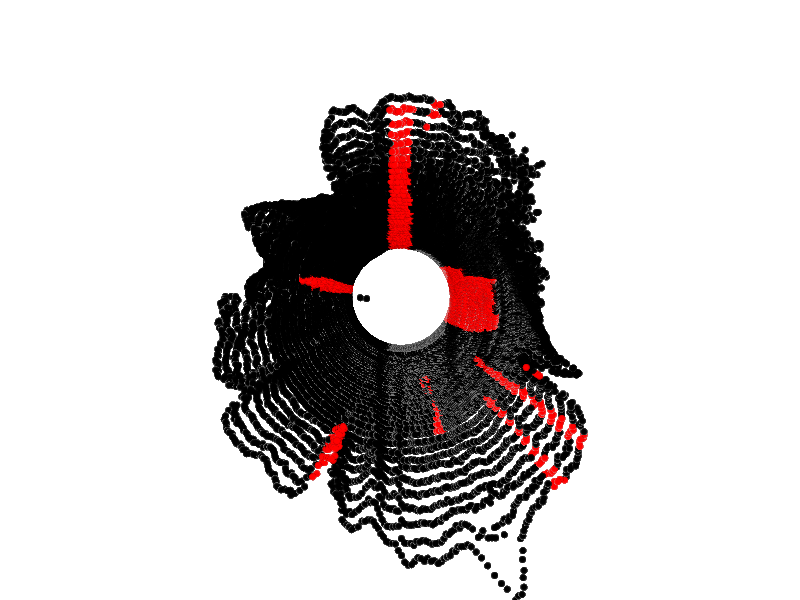}
  \label{fig:6}
\end{subfigure}\hfil
\begin{subfigure}[t]{0.115\textwidth}
  \includegraphics[trim=180 20 80 20,clip,width=\linewidth]{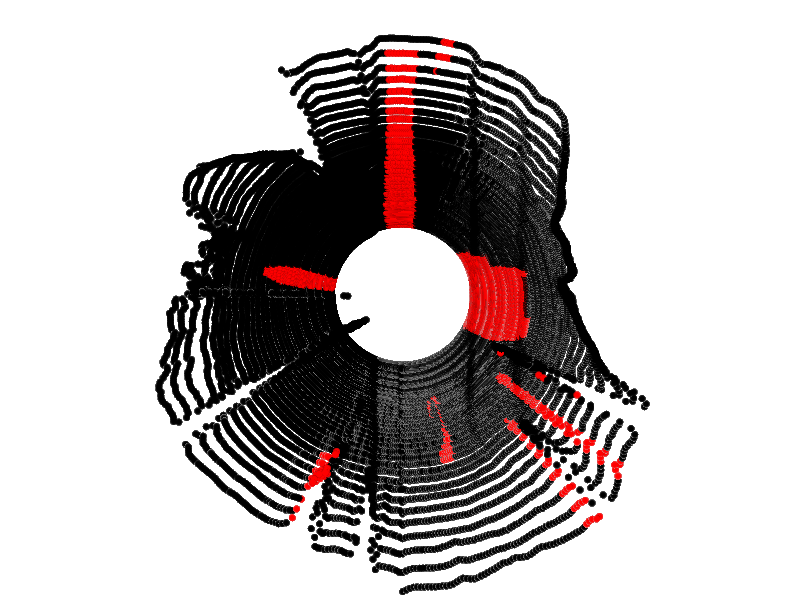}
  \label{fig:3}
\end{subfigure}\hfil
\begin{subfigure}[t]{0.1\textwidth}
  \includegraphics[trim=120 20 0 20,clip,width=\linewidth]{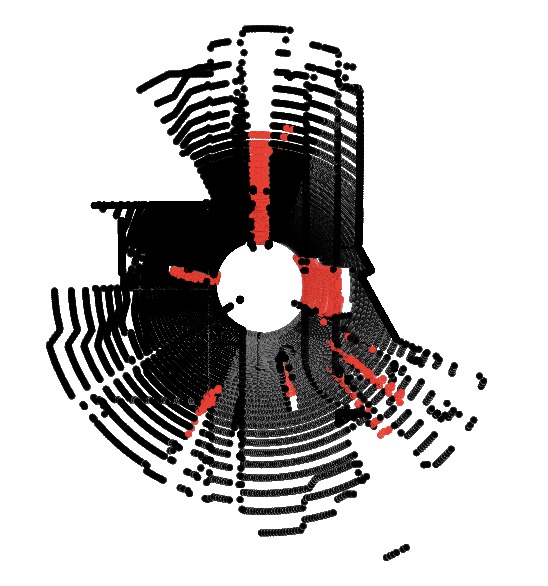}
  \label{fig:20}
\end{subfigure}\hfil
\begin{subfigure}[t]{0.125\textwidth}
  \includegraphics[trim=80 20 20 20,clip,width=\linewidth]{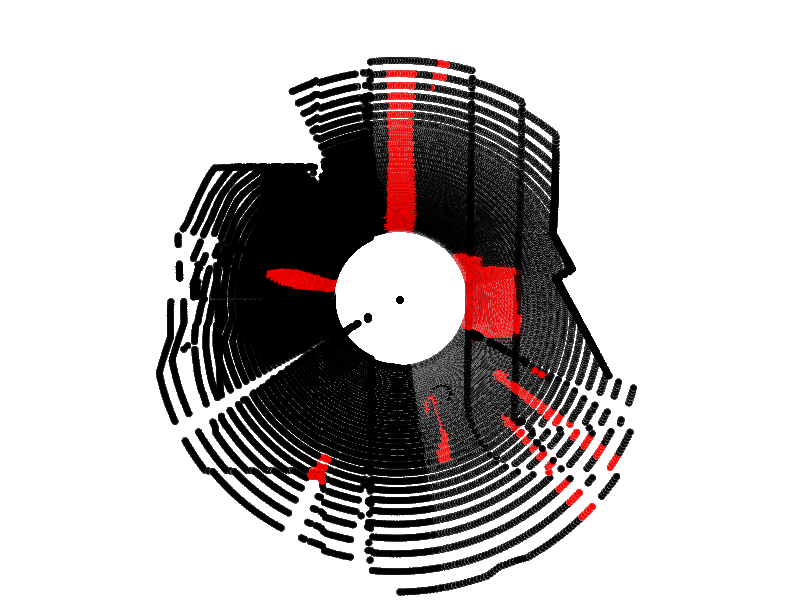}
  \label{fig:2}
\end{subfigure} 

\begin{subfigure}[t]{0.1\textwidth}
\includegraphics[width=\linewidth]{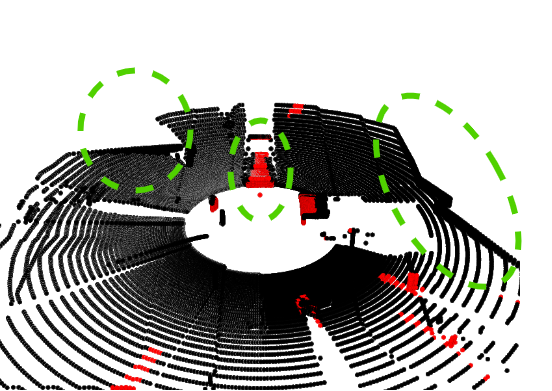}
\captionsetup{font={footnotesize}}
\caption{Dynamic}
\label{fig:7}
\end{subfigure}\hfil
\begin{subfigure}[t]{0.1\textwidth}
\includegraphics[width=\linewidth]{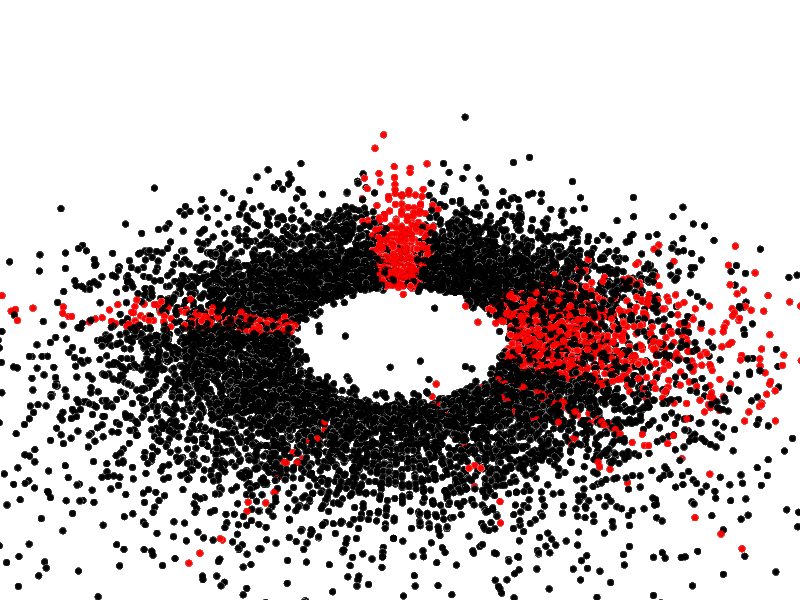}
\captionsetup{font={footnotesize,it}}
\caption{ADMG}
\label{fig:12}
\end{subfigure}\hfil
\begin{subfigure}[t]{0.1\textwidth}
\includegraphics[width=\linewidth]{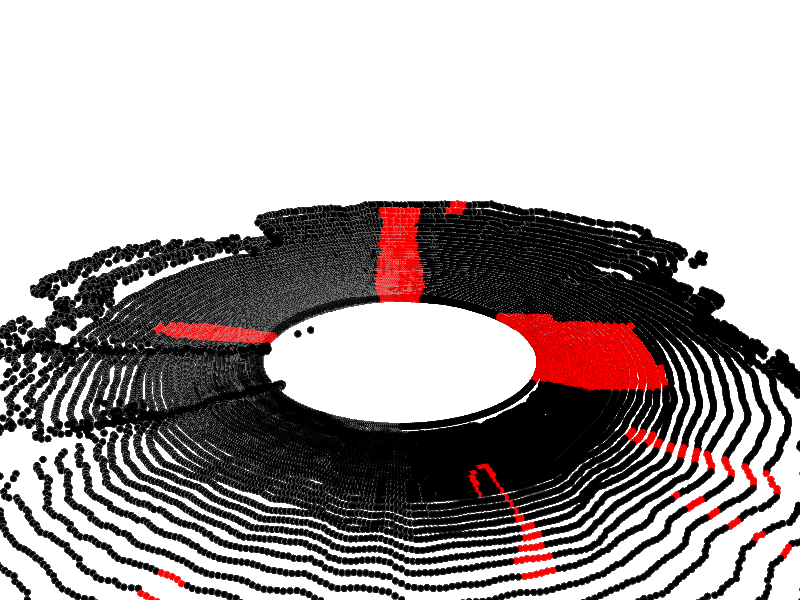}
\captionsetup{font={footnotesize,it}}
\caption{CHCP-AE}
\label{fig:10}
\end{subfigure}\hfil 
\begin{subfigure}[t]{0.1\textwidth}
 \includegraphics[width=\linewidth]{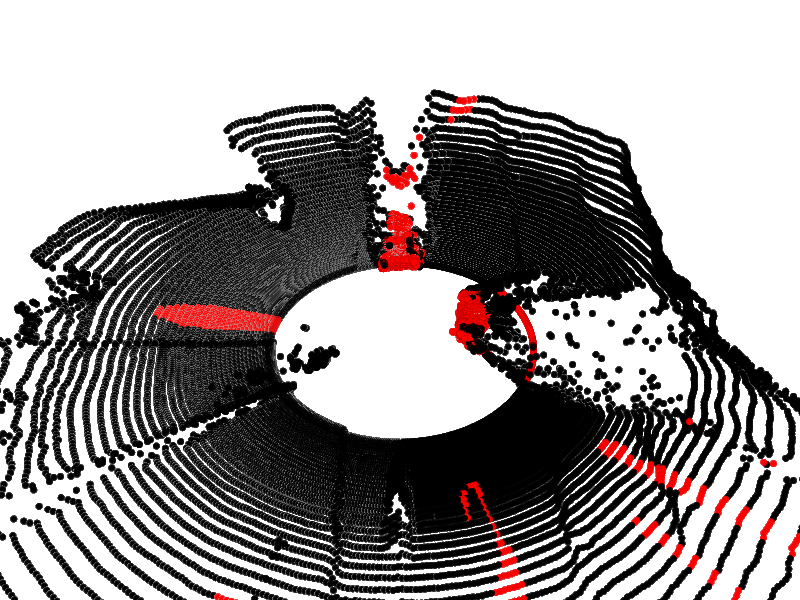}
 \captionsetup{font={footnotesize,it}}
\caption{CHCP-VAE}
\label{fig:11}
\end{subfigure}\hfil 
\begin{subfigure}[t]{0.1\textwidth}
 \includegraphics[width=\linewidth]{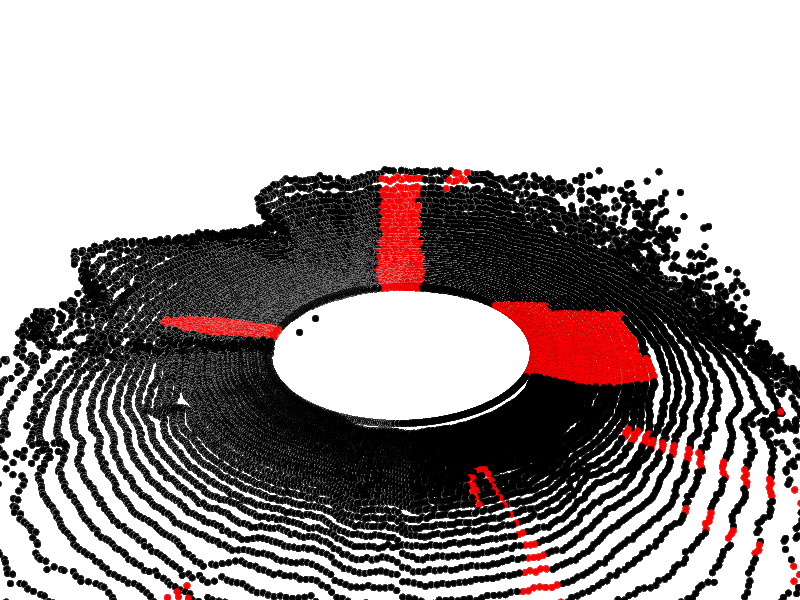}
 \captionsetup{font={footnotesize,it}}
\caption{CHCP-GAN}
\label{fig:12}
\end{subfigure}\hfil
\begin{subfigure}[t]{0.1\textwidth}
\includegraphics[width=\linewidth]{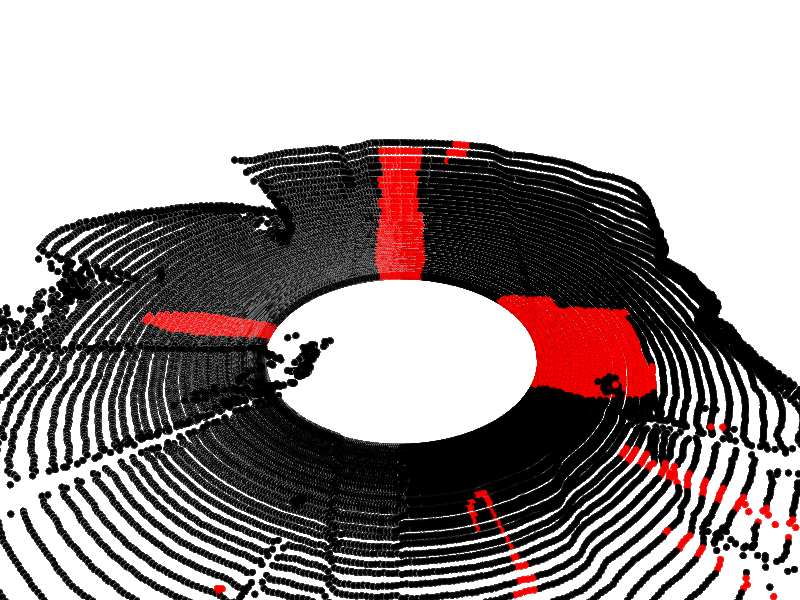}
\captionsetup{font={footnotesize}}
\caption{DSLR}
\label{fig:9}
\end{subfigure}\hfil 
\begin{subfigure}[t]{0.1\textwidth}
\includegraphics[width=\linewidth]{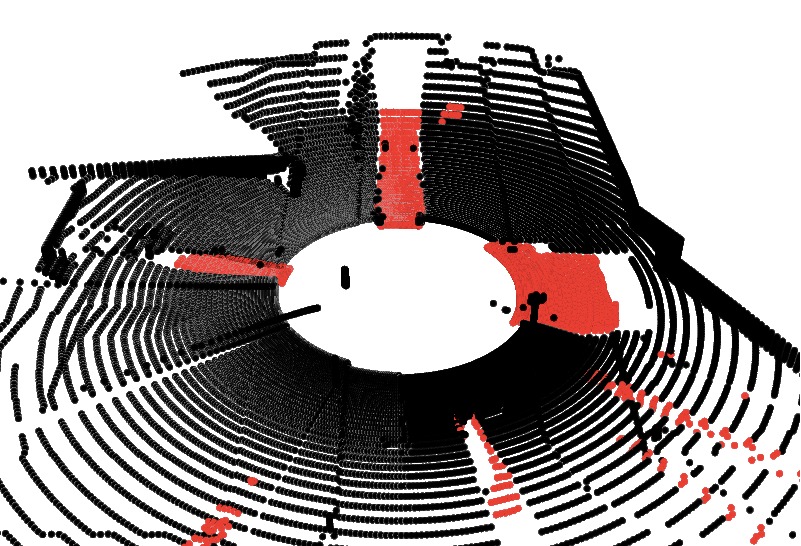}
\captionsetup{font={footnotesize}}
\caption{DSLR-Seg}
\label{fig:12}
\end{subfigure}\hfil
\begin{subfigure}[t]{0.12\textwidth}
  \includegraphics[width=\linewidth]{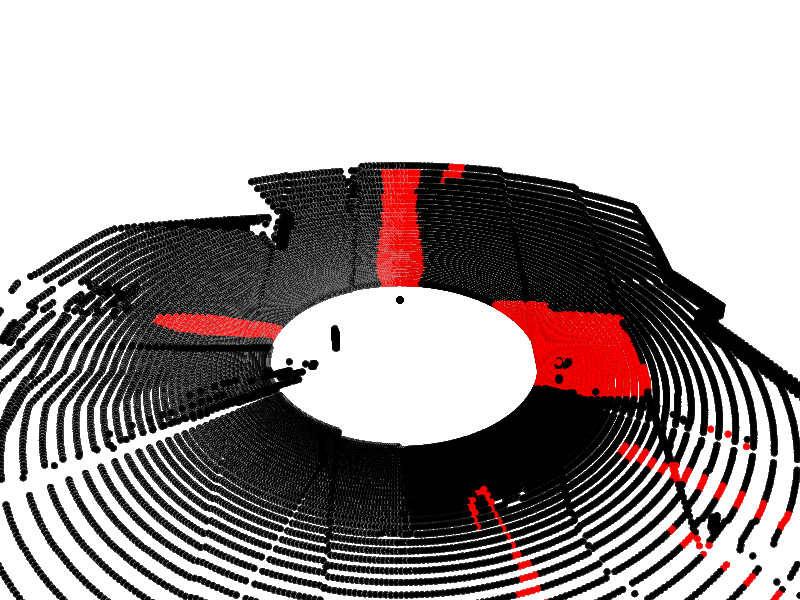}
  \captionsetup{font={footnotesize}}
  \caption{GT-static}
  \label{fig:2}
\end{subfigure} 

\caption{First row shows top-view and bottom row shows lateral view. Reconstruction obtained by DSLR and baseline models on a CARLA-64 scan. Black denotes the static part of the scene. In (a) red denotes dynamic objects, from (b to g) red denotes the reconstructed static background and in (h) red denotes ground truth static. Reconstruction from our model fills occluded regions as well as maintain walls and edges of surrounding scenes clearly compared to the baseline, none of which perform both tasks better compared to DSLR.}
\label{carlaimages}
\end{figure*}

\subsection{Dataset Generation} \label{train_dataset}
 Set of corresponding static-dynamic LiDAR scan pairs are required, to train \DSLR{}. We propose a novel data-collection framework for obtaining this correspondence. We collect data from 2 runs in the same environment. The first run contains only the static-objects while the second run contains both static and dynamic objects along with the ground-truth poses. Using this, we create pairs of dynamic and its corresponding static-frames. Finally, a relative pose transformation is applied as shown in Figure \ref{dataset_building} so that the static structures in both the scans align. All these operations are performed using Open3D \cite{Zhou2018}.

\section{Experimental Results}
We evaluate our proposed approaches against baselines on three different datasets with the following goals: (1) In section \ref{dst_carla}, we evaluate our proposed approaches with adapted baseline models for the problem of DST for LiDAR, (2) In section \ref{slameval}, we evaluate our proposed approaches \DSLR{}, \DSLRSeg{}, and \DSLRUDA{}, for LiDAR based SLAM.
When segmentation information is available, we further improve the quality of our training dataset by identifying the dynamic-points and replace only those points with corresponding static-points obtained from \DSLR{}, as shown in Eq. (\ref{eq-DSLR-seg}). Based on this, we propose \DSLRpp{}, a \DSLR{} model trained on improved dataset created from CARLA-64.

\subsection{Datasets}
\label{Datasets}

\subsubsection{CARLA-64 dataset:}
We create 64-beam LiDAR dataset with settings similar to Velodyne VLP-64 LiDAR on the CARLA simulator \cite{Dosovitskiy17}. 
Similar settings may also improve the model's generalizability by improving performance on other real-world datasets created using similar LiDAR sensors such as KITTI \cite{geiger2012we}.

\subsubsection{KITTI-64 dataset:}
To show results on a standard real-world dataset, we use the KITTI odometry dataset \cite{geiger2012we}, which contains segmentation information and ground truth poses. 

For KITTI, we cannot build a dynamic-static correspondence training dataset using our generation method due to a dearth of static LiDAR scans (less than 7\%) coupled with the fact that most static-scan poses do not overlap with dynamic-scan poses.

\subsubsection{ARD-16:}
We create ARD-16 (Ati Realworld Dataset), a first of its kind real-world paired correspondence dataset, by applying our dataset generation method on 16-beam VLP-16 Puck LiDAR scans on a slow-moving Unmanned Ground Vehicle. We obtain ground truth poses by using fine resolution brute force scan matching, similar to Cartographer. \cite{hess2016real}.

\begin{figure*}[t]
\begin{subfigure}[t]{0.1\textwidth}
  \includegraphics[trim=80 60 120 20,clip,width=\linewidth]{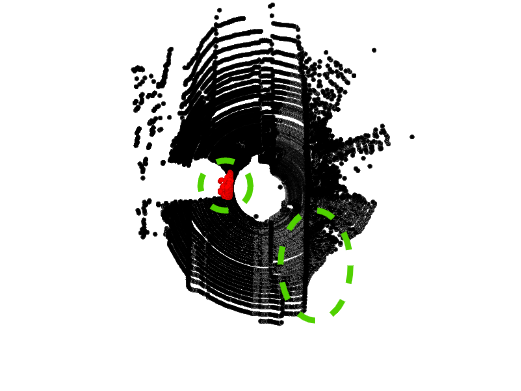}
  \label{fig:1}
\end{subfigure}\hfil 
\begin{subfigure}[t]{0.11\textwidth}
  \includegraphics[trim=200 100 100 80,clip,width=\linewidth]{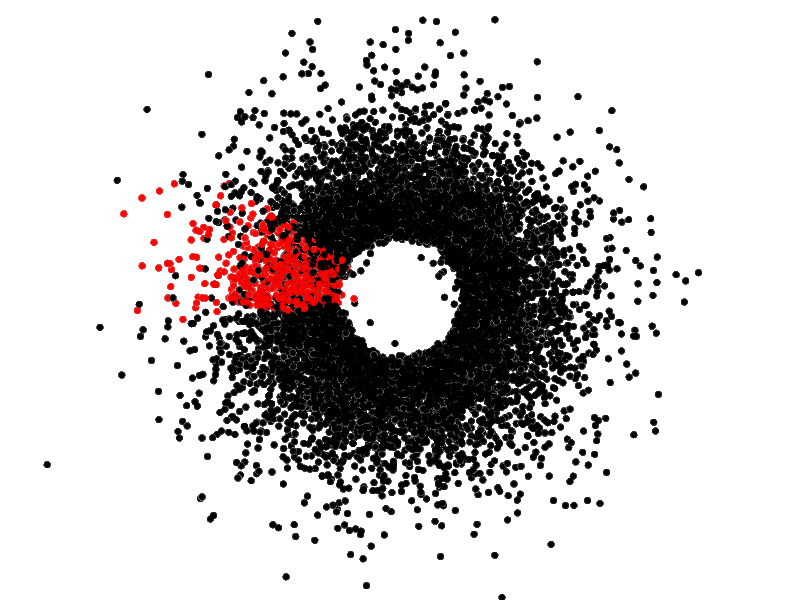}
  \label{fig:19}
\end{subfigure}\hfil
\begin{subfigure}[t]{0.1\textwidth}
  \includegraphics[trim=100 20 120 20,clip,width=\linewidth]{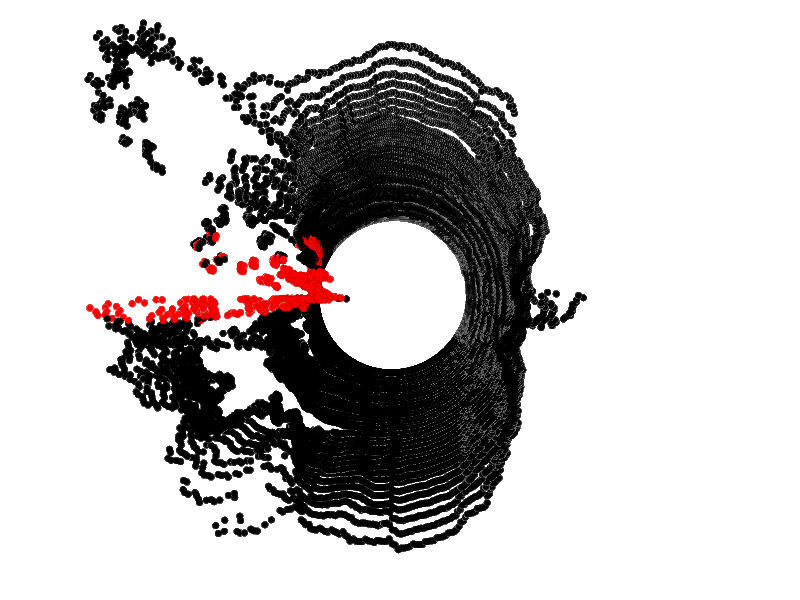}
  \label{fig:4}
\end{subfigure}\hfil
\begin{subfigure}[t]{0.09\textwidth}
  \includegraphics[trim=200 20 100 20,clip,width=\linewidth]{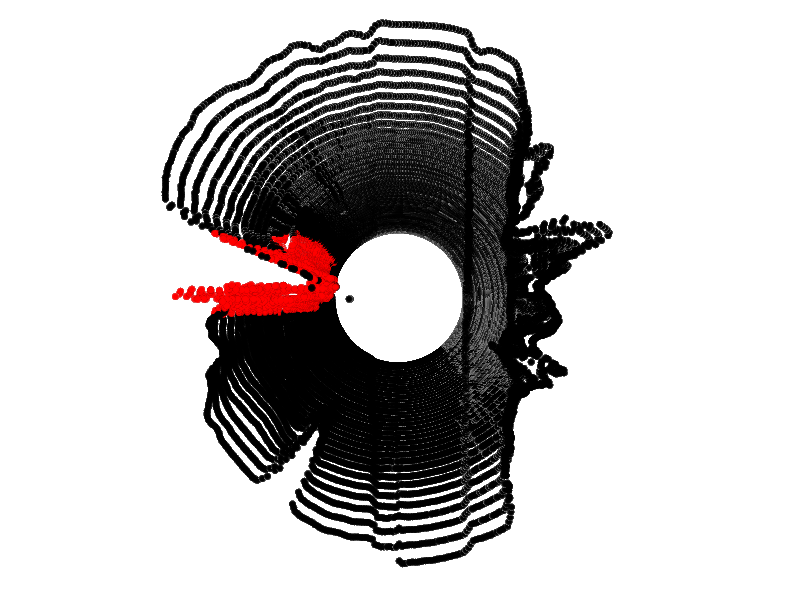}
  \label{fig:5}
\end{subfigure}\hfil
\begin{subfigure}[t]{0.09\textwidth}
  \includegraphics[trim=200 20 100 20,clip,width=\linewidth]{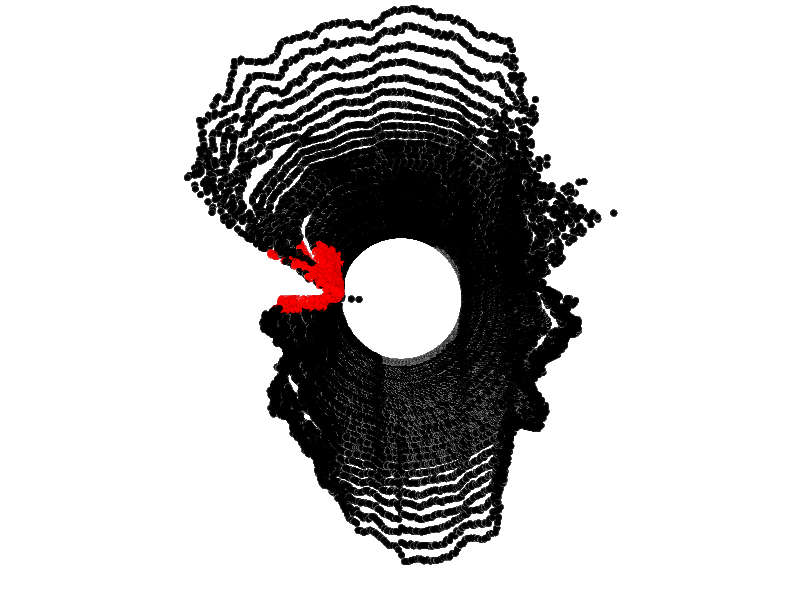}
  \label{fig:6}
\end{subfigure}\hfil
\begin{subfigure}[t]{0.1\textwidth}
  \includegraphics[trim=180 20 20 20,clip,width=\linewidth]{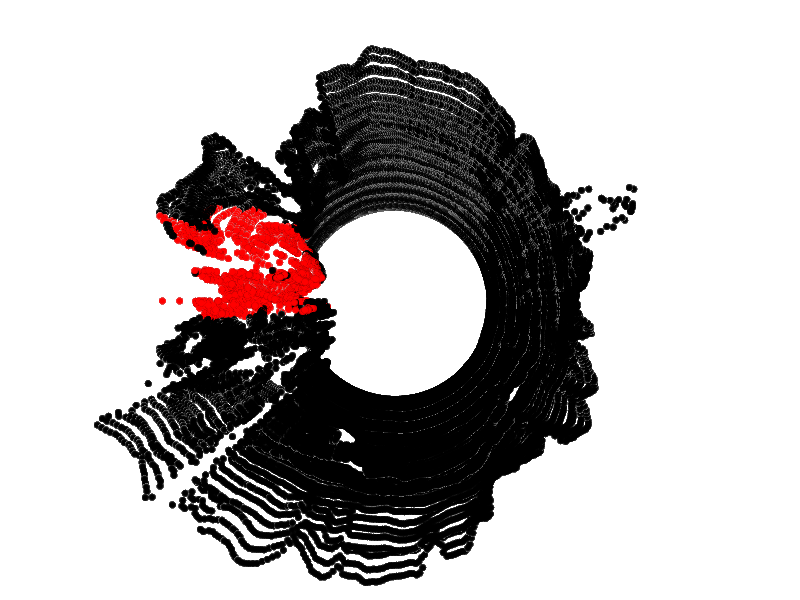}
  \label{fig:20}
\end{subfigure}\hfil
\begin{subfigure}[t]{0.09\textwidth}
  \includegraphics[trim=20 20 20 20,clip,width=\linewidth]{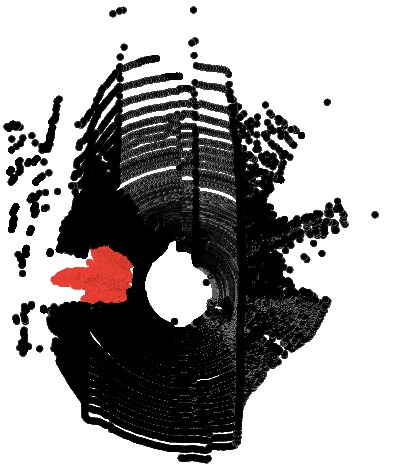}
  \label{fig:200}
\end{subfigure}\hfil
\begin{subfigure}[t]{0.16\textwidth}
  \includegraphics[trim=20 20 20 20,clip,width=\linewidth]{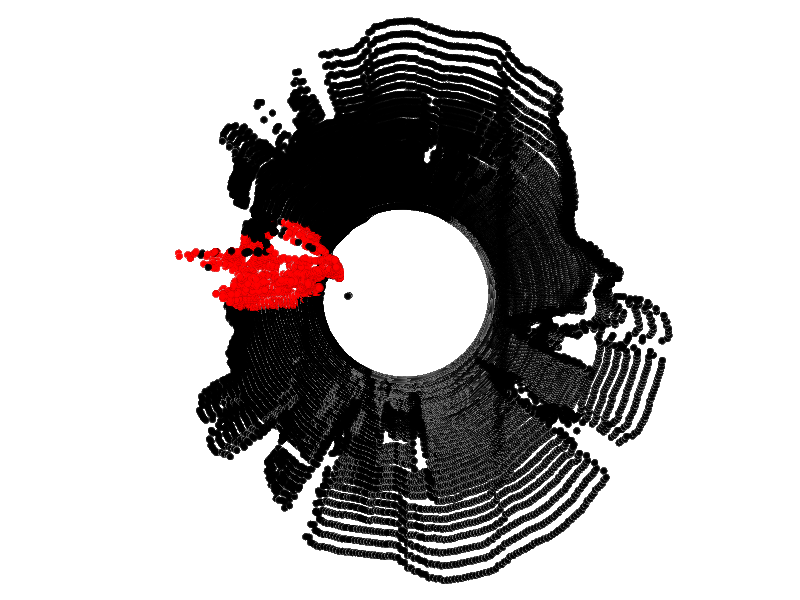}
  \label{fig:3}
\end{subfigure}\hfil

\begin{subfigure}[t]{0.1\textwidth}
\includegraphics[width=\linewidth]{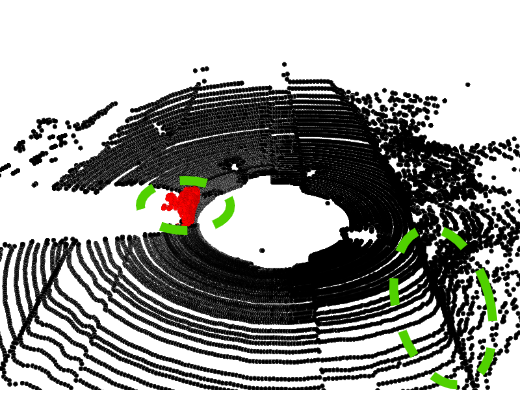}
\captionsetup{font={footnotesize}}
\caption{Dynamic}
\label{fig:7}
\end{subfigure}\hfil 
\begin{subfigure}[t]{0.1\textwidth}
\includegraphics[width=\linewidth]{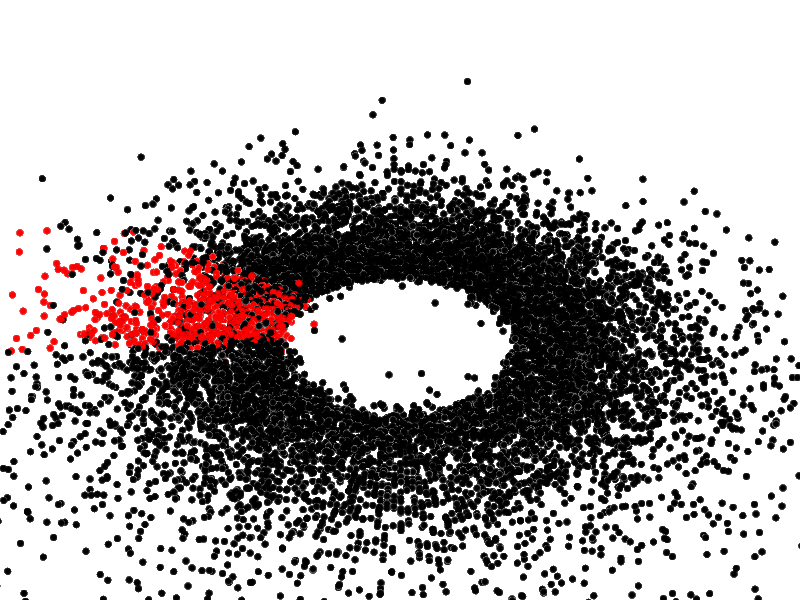}
\captionsetup{font={footnotesize,it}}
\caption{ADMG}
\label{fig:120}
\end{subfigure}\hfil
\begin{subfigure}[t]{0.1\textwidth}
\includegraphics[width=\linewidth]{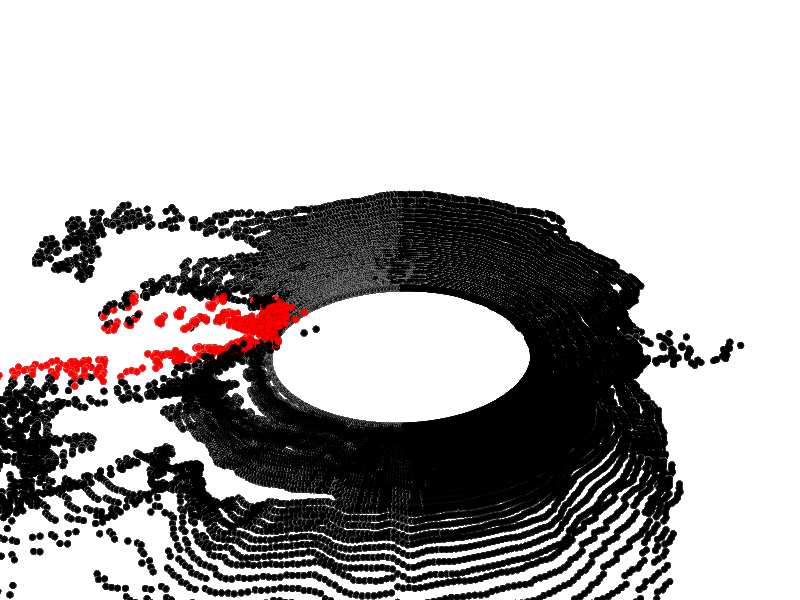}
\captionsetup{font={footnotesize,it}}
\caption{CHCP-AE}
\label{fig:10}
\end{subfigure}\hfil 
\begin{subfigure}[t]{0.1\textwidth}
 \includegraphics[width=\linewidth]{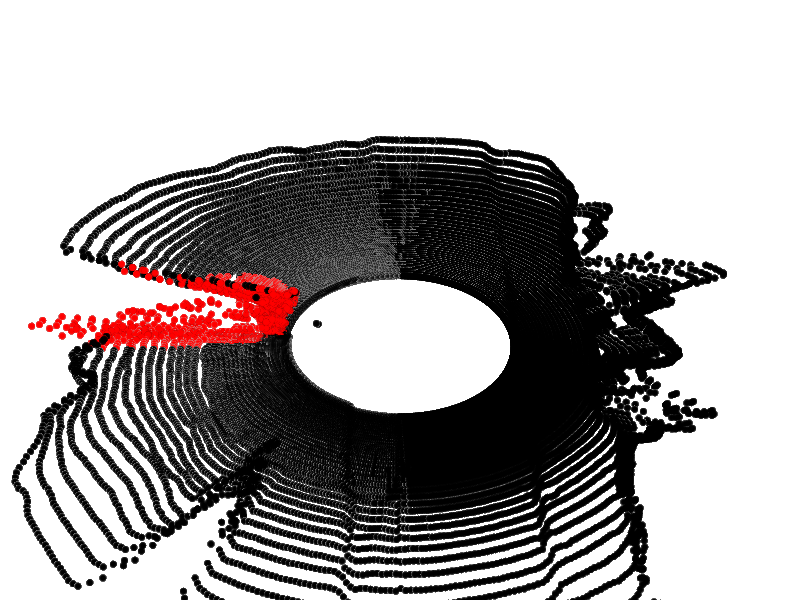}
 \captionsetup{font={footnotesize,it}}
\caption{CHCP-VAE}
\label{fig:11}
\end{subfigure}\hfil 
\begin{subfigure}[t]{0.1\textwidth}
 \includegraphics[width=\linewidth]{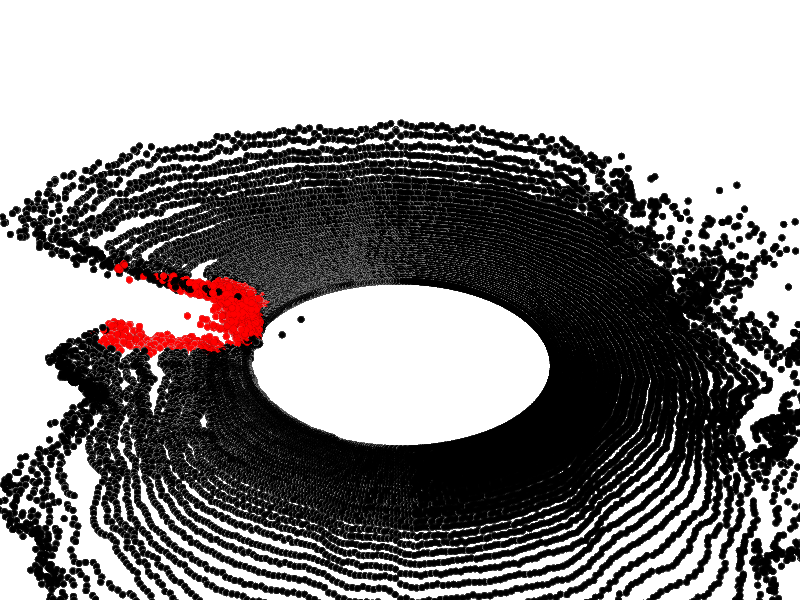}
 \captionsetup{font={footnotesize,it}}
\caption{CHCP-GAN}
\label{fig:12}
\end{subfigure}\hfil
\begin{subfigure}[t]{0.12\textwidth}
\includegraphics[width=\linewidth]{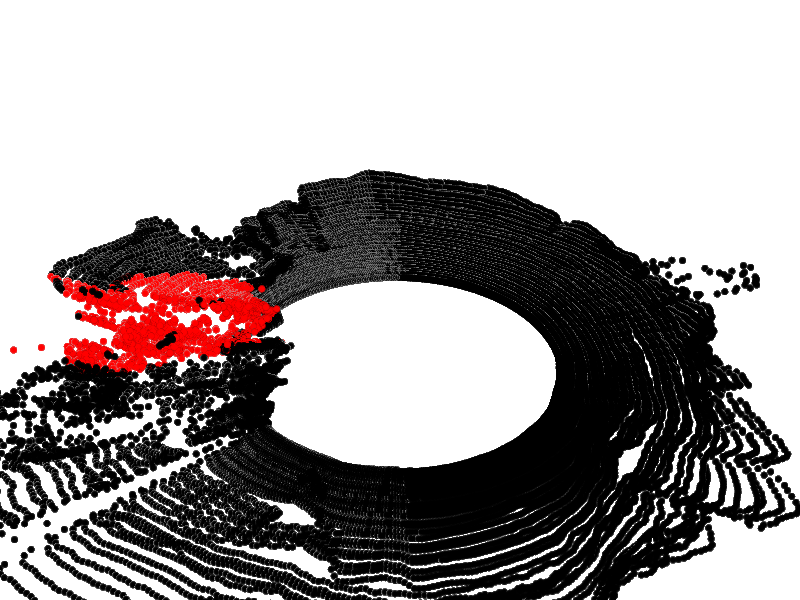}
\captionsetup{font={footnotesize}}
\caption{DSLR}
\label{fig:121}
\end{subfigure}\hfil
\begin{subfigure}[t]{0.1\textwidth}
\includegraphics[width=\linewidth]{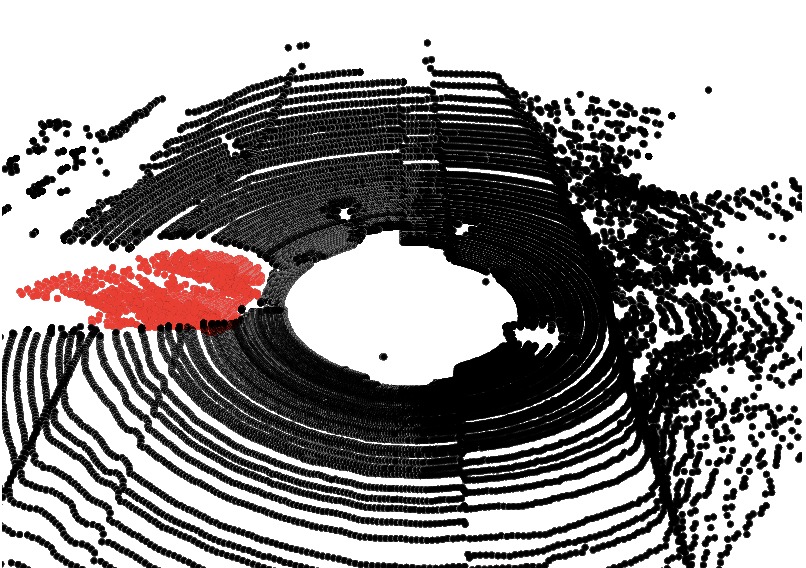}
\captionsetup{font={footnotesize}}
\caption{DSLR-Seg}
\label{fig:125}
\end{subfigure}\hfil
\begin{subfigure}[t]{0.14\textwidth}
\includegraphics[width=\linewidth]{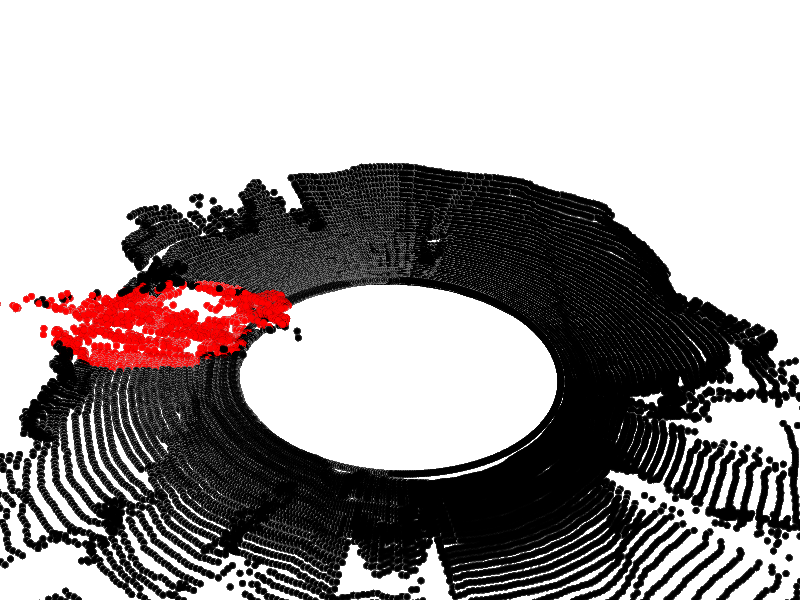}
\captionsetup{font={footnotesize}}
\caption{DSLR-UDA}
\label{fig:9}
\end{subfigure}
\caption{Reconstruction comparison of DSLR along with UDA. Note that KITTI doesn't have the corresponding static. Although CHCP-VAE appears to reconstruct well, it is not able to fill dynamic holes, while DSLR-Seg and DSLR-UDA is able to do both.}
\label{kittiimages}
\end{figure*}

\subsection{Dynamic to Static Translation (DST) on LiDAR Scan}
\subsubsection{Baselines}
\label{Baselines}
 It is established in the existing literature \cite{kendall2015posenet,krull2015learning,bescos2019empty} that learning-based methods significantly outperform non-learning based methods. Hence we adapt the following related work as baselines which leverage learning for reconstruction: (1) \texttt{\textbf{AtlasNet}}\cite{groueix2018papier} (2) \Ach{} \cite{achlioptas2017representation} (3) \CacciaAE{}, \CacciaVAE{}, \CacciaGAN{}\cite{caccia2018deep} (4) \Wu{}\cite{wu2020multimodal} (5) \Bescos{}\cite{bescos2019empty}. For more details on baselines, refer Section \ref{related}.

\subsubsection{Evaluation Metrics}
We evaluate the difference between the model-reconstructed static and ground-truth static LiDAR scans on CARLA-64 and ARD-16 datasets using 2 metrics : Chamfer Distance(CD) and Earth Mover Distance(EMD) \cite{emd-cd}. For more details refer Section 1.4 in Appendix \footnotemark[1].

KITTI-64 does not have corresponding ground truth static, thus we propose a new metric, LiDAR scan Quality Index (LQI) which estimates the quality of reconstructed static LiDAR scans by explicitly regressing the amount of noise in a given scan. This has been adapted from the CNN-IQA model \cite{kang2014convolutional} which aims to predict the quality of images as perceived by humans without access to a reference image. For more details refer Section 1.6 in  Appendix\footnotemark[1].

\subsubsection{DST on CARLA-64:} \label{dst_carla}

We report static LiDAR reconstruction results in Table \ref{table:mainresults} and show that \DSLR{} outperforms all baselines and gives at least 4x improvement on CD over adapted \CacciaAE{}, existing state of the art for DST on LiDAR. Unlike \Wu{} and \Bescos{}, which use segmentation information, \DSLR{} outperforms these without using segmentation information. When segmentation information is available, we extend \DSLR{} to \DSLRpp{} and \DSLRSeg{} which further improves reconstruction performance. For visual comparison, we show the static LiDAR reconstructions in Fig. \ref{carlaimages} generated by our approaches (\DSLR{} and \DSLRSeg{}) and best performing adapted works (\CacciaAE{}, \CacciaVAE{}, \CacciaGAN{}, \Ach{}) on CARLA-64 and KITTI-64. Closer inspection of regions highlighted by dashed green lines shows that \DSLR{} and \DSLRSeg{} accurately in-paint the static background points (shown in red) effectively.

Improved reconstructions using \DSLR{} are due to the imposition of the static structure not only using the reconstruction loss in the image space but also on latent space by forcing the encoder to output latent vector in static domain using discriminator as an adversary.  Therefore compared to \CacciaAE{}, which only uses a simple autoencoder, our use of a discriminator helps identify patterns that help to distinguish the static-dynamic pairs. Also, unlike \Bescos{} \cite{bescos2019empty}, we work directly on the latent space to map dynamic to static point-clouds, rather than working on the pixel/image space.

\subsubsection{DST on KITTI-64:}
\label{dst_kitti}

Due to the unavailability of paired correspondence, models trained on CARLA-64 are tested on KITTI-64 dataset. To adapt to the KITTI-64, we train \DSLR{} using UDA only for \DSLRUDA{}. Since the ground-truth static is not available, we use the proposed LQI to estimate the quality of reconstructed static LiDAR scans and report the results in Table \ref{table:mainresults}. We plot LQI v/s CD in Fig. \ref{ate_vs_cd} and further show that LQI positively correlates with CD on CARLA-64. For more details, refer Table 1 in Appendix\footnotemark[1].

We show that our approaches, \DSLR{}, \DSLRSeg{}, and \DSLRUDA{} outperforms the adapted \CacciaVAE{}, the best performing baseline on KITTI-64 based on LQI. In Table \ref{table:mainresults} and Figure \ref{kittiimages}, we show that although \CacciaVAE{} is the best performing baseline but like all other baselines, it fails to in-paint accurately. On the contrary, our approach tries to retain the static 3D structures as much as possible and in-paint them accurately. \DSLRUDA{} trained on KITTI-64 in an unsupervised manner does a better job at recovering road boundaries in LiDAR scan than \DSLR{}. We can also see that using segmentation information, \DSLRSeg{} is able to perfectly recover static structures like walls and road boundaries.

\begin{table}[t]
\centering
\fontsize{9pt}{\baselineskip}\selectfont
\setlength\extrarowheight{2pt}
\begin{tabular}{c|c|c|c}
\hline
Model & \multicolumn{1}{c}{CARLA-64} & \multicolumn{1}{c}{ARD-16} & KITTI-64  \\\hline
 &  Chamfer & Chamfer & LQI\\
 \cline{2-4} 
AtlasNet  & 5109.98 & 176.46 & -\\
ADMG      & 6.23 & 1.62 & 2.911  \\
CHCP-VAE  & 9.58 & 0.67 & 1.128 \\
CHCP-GAN & 8.19 & 0.38 & 1.133 \\
CHCP-AE & 4.05 & 0.31 & 1.738 \\
\textit{WCZC}  & 478.12 & - & - \\
\textit{EmptyCities}  & 29.39 & - & -  \\
\hline
DSLR (Ours) & \textbf{1.00} & \textbf{0.20} & \textbf{1.120} \\
\textit{DSLR++} (Ours) & \textbf{0.49} & - & - \\
\textit{DSLR-Seg} (Ours) & \textbf{0.02} & -  & - \\
\textit{DSLR-UDA}(Ours) & - & - & 1.119  \\
\hline
\end{tabular} 
\caption{Comparison of LiDAR reconstruction baselines on different datasets. Lower is better. Models that require segmentation information have been italicised for reference.}
\label{table:mainresults}
\end{table}

\begin{table}[h!]
\fontsize{9pt}{\baselineskip}\selectfont

\begin{tabular}{ccccc}
\hline\noalign{\smallskip}
 Model & ATE & Drift & \multicolumn{2}{c}{RPE}\\ \hline
  &       &       &      Trans & Rot \\
 \cline{4-5} 
\noalign{\smallskip}

 \multicolumn{5}{c}{KITTI-64 Dataset} \\
\noalign{\smallskip}
 Pure-Dynamic &  11.809 & 14.970 & 1.620 & 1.290\textdegree \\ 
  Detect \& Delete (MS) & 12.490 & \textbf{13.599} & 1.623 & 1.290\textdegree  \\ 
  Detect \& Delete (GTS) & \textbf{11.458} & 22.856 & 1.630 & 1.336\textdegree  \\ 
   DSLR-Seg (MS)  & 99.642 & 34.372 & \textbf{1.610} &  \textbf{1.290}\textdegree \\
  DSLR-Seg (GTS)  & 11.699 & 19.67 & 1.620 &  1.295\textdegree \\
\hline
\noalign{\smallskip}
\multicolumn{5}{c}{CARLA-64 Dataset} \\
\noalign{\smallskip}
 Pure-Dynamic &  10.360 & 18.580 & 0.056 & 0.403\textdegree \\ 

  Detect \& Delete (MS) & 10.430 & 18.870 & 0.060 & 0.430\textdegree  \\ 
  Detect \& Delete (GTS) & 11.650 & 24.430 & 0.063 & 0.401\textdegree  \\ 
   DSLR-Seg (MS)  & 10.760 & 17.430 & 0.050 &  0.390\textdegree \\
  DSLR-Seg (GTS)  & \textbf{7.330} & \textbf{13.63} & \textbf{0.050} &  \textbf{0.340}\textdegree \\
\hline
\noalign{\smallskip}
\multicolumn{5}{c}{ARD-16 Dataset} \\
 \noalign{\smallskip}
 Pure-Dynamic &  1.701 & -- & 0.036 & \textbf{0.613}\textdegree \\ 
  DSLR (Ours)  & \textbf{1.680} & -- & \textbf{0.035} &  0.614\textdegree \\
\hline

\end{tabular}
\caption{Comparison of SLAM  performance with existing Baselines.}
\label{table:slamdataset}
\end{table}

\subsubsection{DST on ARD-16:}
In Table \ref{table:mainresults}, we also report static LiDAR scan reconstruction results on ARD-16 dataset. We do not report results on \Bescos{}, \Wu{}, \DSLRpp{}, \DSLRSeg{} due to the unavailability of segmentation information. To the best of the author's knowledge, we are the first to train and test a DST based model on real-world data owing to the dataset generation pipeline described in Section \ref{train_dataset}. We show that \DSLR{} outperforms adapted baselines on all metrics on this dataset and demonstrate the practical usefulness of our method in real-world setting.

We summarize the results of \DSLR{} on the 3 datasets here. We conclude that \DSLR{} gives significant improvement over the best performing baseline model \CacciaAE{} over both metrics. Moreover, if segmentation information is  available, our variants \DSLRpp{} and \DSLRSeg{} further extend the gains as shown in Table \ref{table:mainresults}. We also show the efficacy of our model on a real world dataset, ARD-16, where our model performs significantly better. We do not have segmentation information available for ARD-16. If segmentation information is available for the real world dataset, the gains can be higher. For a detailed comparison of LiDAR reconstruction baselines  refer Table 1 in Appendix\footnotemark[1].

\subsection{Application of LiDAR Scan Reconstruction for SLAM in Dynamic Environments}
\label{slameval}

 We use Cartographer \cite{hess2016real}, a LiDAR based SLAM  algorithm \cite{filipenko2018comparison,yagfarov2018map} to test the efficacy of our proposed approaches \DSLR{} and \DSLRSeg{} on tasks pertaining to autonomous navigation. Our reconstructed output serves as the input LIDAR scan for Cartographer, which finally gives a registered map of the environment and  position of the vehicle in real-time.

\subsubsection{Baselines:}
We evaluate our model with two baselines: (1) A unprocessed dynamic frame (\textbf{Pure-Dynamic}) and (2) With an efficient LiDAR pre-processing step for handling SLAM in dynamic environment which removes the dynamic points from the LiDAR scan (\textbf{Detect \& Delete})  \cite{ruchti2018mapping,vaquero2019improving}.

\subsubsection{Evaluation Metrics:}
After coupling the Cartographer with above prepossessing steps, the generated output (position of the vehicle) is compared on widely used SLAM  metrics like Absolute Trajectory Error (ATE) \cite{6385773}, Relative Position Error (RPE) \cite{6385773} and Drift. For more details refer  Section 3 in Appendix\footnotemark[1].

\begin{figure}[t]

    \centering
    \includegraphics[scale=0.25]{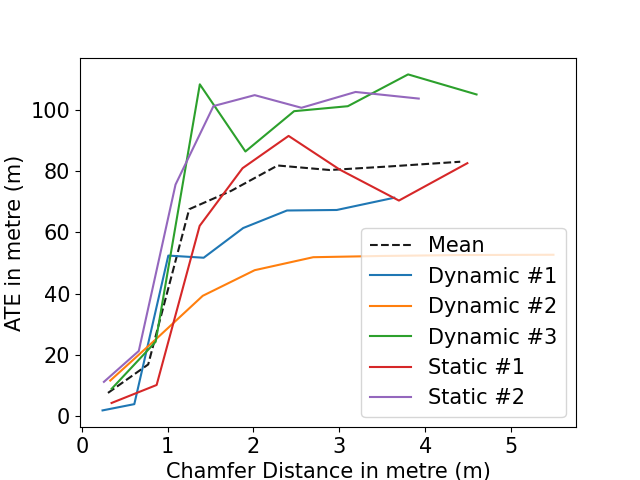}
    \includegraphics[scale=0.25]{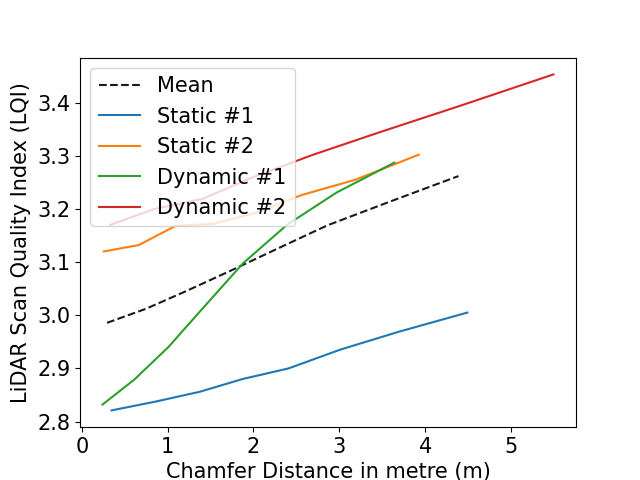}
    \caption{Studying effect of different metrics on Chamfer distance. Left: ATE vs CD plot 
    Right: LQI vs CD plot
    }
    \label{ate_vs_cd}
\end{figure}

\begin{figure}[h]
    \centering
    \includegraphics[scale=0.25]{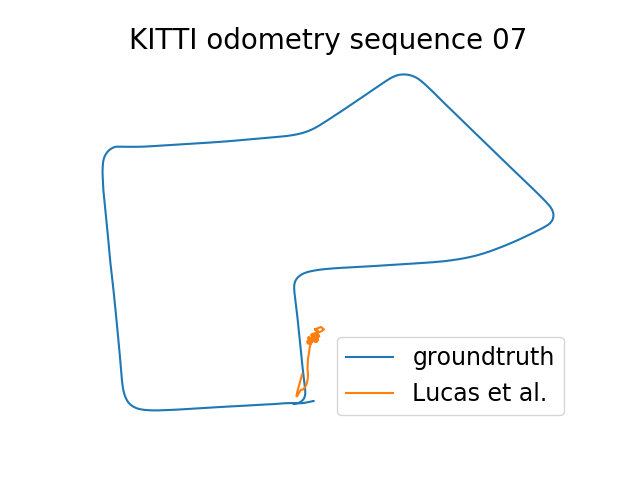}
    \includegraphics[scale=0.25]{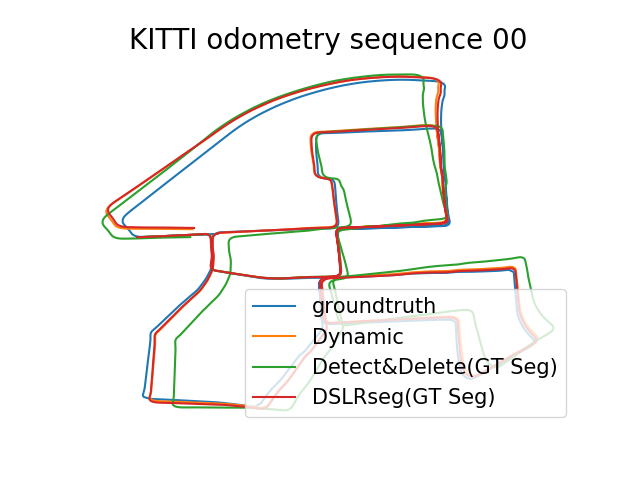}
    \caption{(1) Left: CHCP-AE baseline with ground truth evaluation on KITTI (2) Right: Detect \& Delete baseline and DSLR-Seg with Model-Seg variation on KITTI.
    }
    \label{slamtraject}
\end{figure}

\subsubsection{Discussion}
\label{slam_baseline}

 We compare our best performing model in each dataset against dynamic SLAM baseline methods.  We report results with two variations of segmentation for CARLA-64 and KITTI-64 to limit the effect of segmentation inaccuracy in our interpretation. We report MS variant when U-Net was used to give segmentation masks for dynamic points and GTS variation when ground truth segmentation masks were used.  In Table \ref{table:slamdataset}, we show that our approaches outperform the base case Pure-Dynamic in most of the cases and perform competitively against dynamic SLAM baseline.  We choose \DSLRSeg{} for comparison on CARLA-64 and KITTI-64 and \DSLR{} for comparison on ARD-16 due to the unavailability of segmentation information for ARD-16. For a detailed analysis refer to Section 3 of the Appendix\footnotemark[1]. 

We also study the relation between LiDAR reconstruction error (CD) and SLAM error (ATE). For this, we take random dynamic and static scans from CARLA-64 dataset and linearly increase LiDAR reconstruction error by corrupting them with Gaussian noise and report SLAM errors for the same. In Figure \ref{ate_vs_cd}, we observe that for most runs, SLAM error initially increases exponentially with linear increase in LiDAR error and then saturates at a CD of 2m. Based on this, we conclude that, for any reconstruction model to be practical for SLAM, it must have reconstruction error or CD below 2m which we refer to it as SLAM Reconstruction Threshold (SRT). We see from CARLA-64 results in Table \ref{table:mainresults} that \DSLR{} is the only model to have its reconstruction error below the SRT and therefore is practically feasible for SLAM. We show this experimentally by plotting the trajectory estimated by \CacciaVAE{}, the best performing static LiDAR scan reconstruction baseline in Figure \ref{slamtraject} and show that pose estimated by Cartographer gets stuck after some time due to high LiDAR reconstruction error, making it impractical.Through this study we establish the superiority of our proposed models over existing approaches in static LiDAR scan reconstruction for SLAM.

\section{Conclusion}
We propose a novel, adversarially-trained, autoencoder model for static background LiDAR reconstruction without the need of segmentation masks. We also propose variants that accept available segmentation masks for further improving LiDAR reconstruction. It also enhances SLAM performance in dynamic environments. However some drawbacks in our reconstruction are discussed in Appendix\footnotemark[1]. Future work to address those issues can help improve the reconstruction quality even more and their applicability for tasks in real world environments.

\section*{Acknowledgements}
The authors thank Ati Motors for their generous financial support in conducting this research.

\bibliography{egbib}

\end{document}


\maketitle


\section{Implementation Details}

\subsection{LiDAR scan to Range image}

\begin{figure}[h!]
{%
  \includegraphics[clip,width=\columnwidth]{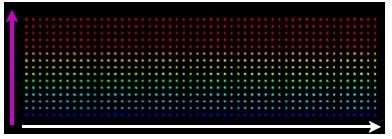}%
}
\caption{Image taken from \cite{caccia2018deep}.Points at the same elevation angle have the same color. Every row denotes a 360\textdegree scan of the environment at a certain elevation angle. Every column denotes points at the same azimuth angle for different elevation angles.}
\label{fig:grid}
\end{figure}

The LiDAR scan obtained from VLP-64 LiDAR gives $\approx$ 1,30,000 points per scan. Geometrically, the scan consists of 64 laser beams having rotated 360\textdegree to give the point cloud its structure. We divide the azimuth angle into 512 bins while the elevation angle in 64 bins (due to 64 beams). This gives a grid structure of 64 $\times$ 512 where every cell consists of \textit{x,y,z} coordinate calculated by averaging all points falling in the cell. It is to be noted that we can avoid the averaging by setting the number of bins for the azimuth angle equal to the number of points given by one laser beam in a 360\textdegree scan. We do not do this due to computational limitations. As observed in figure \ref{fig:grid} from \cite{caccia2018deep}, points far away from the origin for a scan are most noisy and are most difficult to train and thereby are removed. For more details refer \cite{caccia2018deep}.

\subsection{Model architecture}

\subsubsection{Autoencoder}
We use an autoencoder inspired from \cite{caccia2018deep}. The network uses a DCGAN based discriminator-generator as encoder-decoder. The input is LiDAR based range-image which is reconstructed at the output. The training procedure is shown in the Algorithm\ref{AutoencoderAlgo}.

\begin{algorithm}[t]
\SetAlgoLined
 $E(\phi)$ : DCGAN discriminator based Encoder \\
 $G(\theta)$ : DCGAN generator based Decoder \\
 $L$ : LiDAR frame based images \\
 \For{$x \in L$}
 {
    $AELoss = 0$ \\
    $x \xrightarrow[]{E(\phi)} r(x)$ \\
    $r(x) \xrightarrow[]{G(\theta)} \overline{x}$ \\
    $AELoss = MSE(x, \overline{x})$ \\
     Train $\phi, \theta$ using AELoss
 }
 \caption{AUTOENCODER}
 \label{AutoencoderAlgo}
\end{algorithm}

\subsubsection{Pair Discriminator}
Latent embeddings of a given LiDAR scan are obtained via an autoencoder. In a straightforward approach, one could use the above autoencoder and learn the proposed paired discriminator, that differentiates (static, static) vs (static, dynamic) LiDAR embedding pairs, by using the latent embedding given by it for static and dynamic frames. However, given that dynamic and static LiDAR scans look alike, except for the occlusions due to dynamic objects. The embedding space of both have a high overlap and simply using the above technique for training the paired discriminator doesn't perform well.

Let $E_{S}$ and $E_{D}$ be embedding space for static and dynamic LiDAR scans respectively. We hypothesize that \textit{($E_{S} \cup E_{D}) - (E_{S} \cap E_{D}$)} differentiates static and dynamic scans. In other words, segmentation information can help differentiate the scans. However, as we do not use the segmentation information, we experimentally find that fixing the LiDAR latent space and training the discriminator doesn't yield the desired result. Letting the latent space change as the discriminator training progresses i.e., also training the autoencoder using reconstruction loss, helps the paired discriminator yield desired results. Therefore, without even using segmentation information, the discriminator can learn to focus on the regions of difference (occluded regions in dynamic frames) and is able to differentiate between (static, static) and (static, dynamic) LiDAR latent embedding pairs. This discriminator when used in adversarial training helps to map an embedding in dynamic latent-space to its corresponding static embedding in static latent-space. The training procedure for the discriminator is shown in algorithm \ref{DiscriminatorAlgo}.

\begin{algorithm}[t]
\SetAlgoLined
 $DI(\gamma)$ : Discriminator \\
 $S$ : Static LiDAR Images \\
 $D$ : Dynamic LiDAR Images \\
$G(\phi,\theta)$ : Autoencoder for Static Frames \\
 \For{$i \in range( 0, |S|)$}
 {
    \For{$j \in range( 0, |S|)$}
    {
        $Dual Loss = Loss DI = Loss AE = 0$ \\
        $S_{i} \xrightarrow[]{\phi} r(S_{i}) \xrightarrow[]{\theta} \overline{S_{i}}$ \\
        $S_{j} \xrightarrow[]{\phi} r(S_{j}) \xrightarrow[]{\theta} \overline{S_{j}}$ \\
        $D_{j} \xrightarrow[]{\phi} r(D_{j}) \xrightarrow[]{\theta} \overline{D_{j}}$ \\
        $DI( r(S_{i}), r(S_{j})) \xrightarrow[]{} d_{SS} \in [0, 1]$ \\
        $Loss DI = BCE(D_{SS}, 1)$ \\
        $Loss AE = MSE(\overline{S_{i}}, S_{i}) + MSE(\overline{S_{j}}, S_{j})$ \\
        $Dual Loss = \alpha * Loss DI + Loss AE $ \\
        Backprop $Dual Loss$ to train $\phi, \theta, \gamma$ \\
        $Dual Loss = Loss DI = Loss AE = 0$ \\
        $DI( r(S_{i}), r(S_{j})) \xrightarrow[]{} d_{SD} \in [0, 1]$ \\
        $Loss DI = BCE(D_{SD}, 0)$ \\
        $Loss AE = MSE(\overline{D_{j}}, D_{j})$ \\
        $Dual Loss = \alpha * Loss DI + Loss AE $ \\
        Backprop $Dual Loss$ to train $\phi, \theta, \gamma$
    }
 }
 \caption{Discriminator}
 \label{DiscriminatorAlgo}
\end{algorithm}

\subsubsection{Adversarial Training}
\begin{algorithm}[h]
\SetAlgoLined
 $G_{static}(\phi_{SE}, \theta_{SD})$ : Autoencoder taking static input \\
 $G_{dynamic}(\phi_{DE}, \theta_{DD})$ : Autoencoder taking dynamic input \\
 $DI(\gamma)$ : Discriminator \\
 $\phi_{SE}, \theta_{SD}, \theta_{DD}, \gamma$ : non-trainable weights \\
 $\phi_{DE}$ : trainable weights \\
 $D$ : {Set of Dynamic Frames} \\
 $S$ : {Set of corresponding Static Frames} \\
 \For{$i \in range( 0, |S|)$}
 {
    \For{$j \in range( 0, |S|)$}
    {
        $S_{i} \xrightarrow[]{\phi_{SE}} r(S_{i}) \xrightarrow[]{\theta_{SD}} \overline{S_{i}}$ \\
        $D_{j} \xrightarrow[]{\phi_{DE}} r(D_{j}) \xrightarrow[]{\theta_{DD}} \overline{D_{j}}$ \\
        $DI( S_{i}, D_{fj}) \xrightarrow[]{} d_{SD} \in [0, 1]$ \\
        $Adv Loss DI = BCE(d_{SD}, 1)$ \\
        $Loss AE = MSE(\overline{D_{j}}, S_{i})$ \\
        $Adv Dual Loss= \alpha*Adv Loss DI +LossAE $ \\
         Train $\phi_{DE}$ using $Adv Dual Loss$ \\
    }
 }
 \caption{Adversarial Training}
 \label{AdversarialAlgo}
\end{algorithm}

\begin{algorithm}[b!]
\SetAlgoLined
    $D$ : a random dynamic run is a set of dynamic LiDAR scans $d_{i}$ in at pose $p_{i}$\\
    $S$ : a random static run is a set of static LiDAR scans $s_{j}$ at pose $q_{j}$\\
    $T_{r}(s_{j})$ : LiDAR scan $s_{j}$ transformed rigidly based on pose $r$ \\
    $\triangle (p_{i}, q_{j})$ : relative pose difference between two LiDAR scans $d_{i}$ and $s_{j}$ \\
    $\delta$ : desirable/given relative pose difference threshold for paired correspondence matching \\
    $M$ : training dataset for \DSLR{} is a set of matching paired correspondence pair\\
    \textbf{Input:} Run $S$ and Run $D$ \\
    \textbf{Output:} $M$ \\
    $M \xleftarrow[]{} \phi$ \\
    \For{$(d_{i}, p_{i}) \in D$}
    {
        \For{$(s_{j}, q_{j}) \in S$}
        {
            $r \xleftarrow[]{} \triangle (p_{i}, q_{j})$ \\
            \If{$ r < \delta$}
            {
                $M \xleftarrow[]{} M + (d_{i}, T_{r}(s_{j}))$ \\
            }
        }
    }
    \caption{Paired Correspondence Dataset Generation}
    \label{corr_dataset_algo}
\end{algorithm}

\begin{figure*}[t!]
\begin{subfigure}[t]{0.175\textwidth}
\centering
  \fbox{\includegraphics[width=3cm,height=2cm]{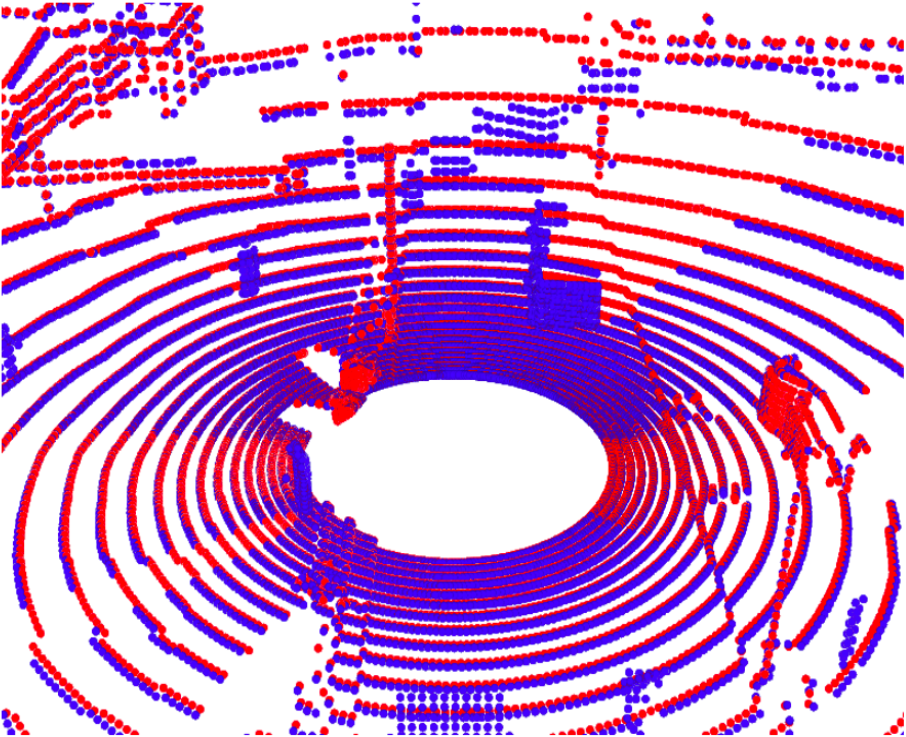}}
\end{subfigure}\hfil 
\begin{subfigure}[t]{0.175\textwidth}
\centering
  \fbox{\includegraphics[width=3cm,height=2cm]{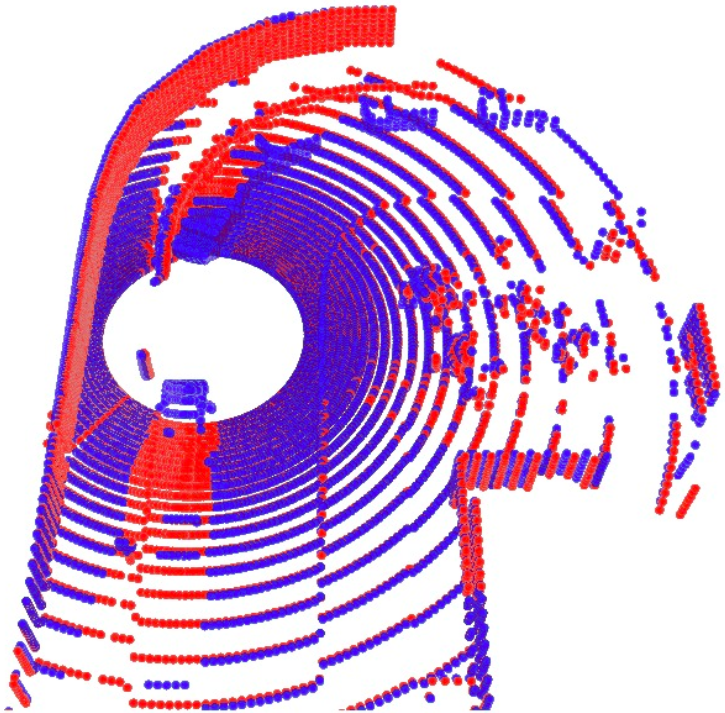}}
\end{subfigure}\hfil
\begin{subfigure}[t]{0.175\textwidth}
\centering
  \fbox{\includegraphics[width=3cm,height=2cm]{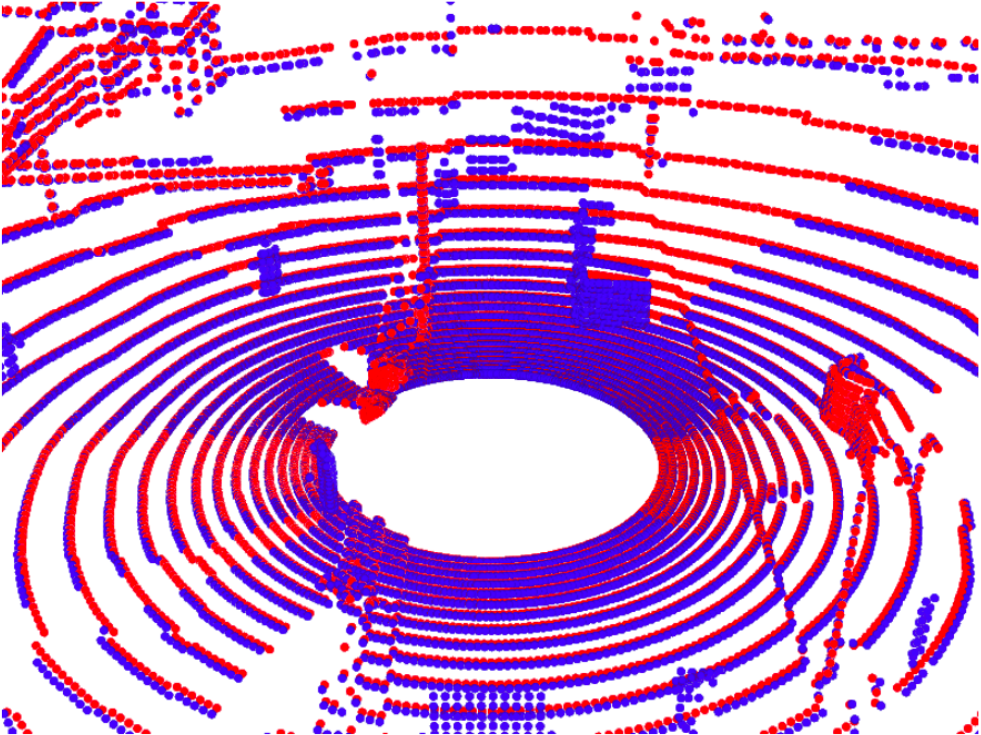}}
\end{subfigure}\hfil
\begin{subfigure}[t]{0.175\textwidth}
\centering
  \fbox{\includegraphics[width=3cm,height=2cm]{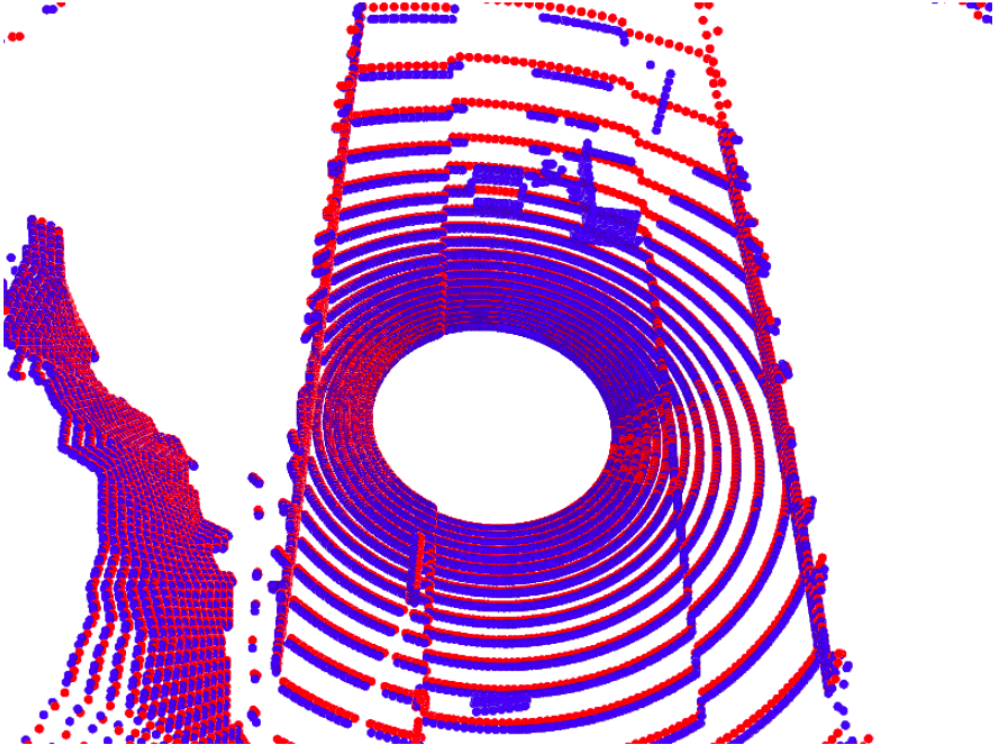}}
\end{subfigure}\hfil
\begin{subfigure}[t]{0.175\textwidth}
\centering
  \fbox{\includegraphics[width=3cm,height=2cm]{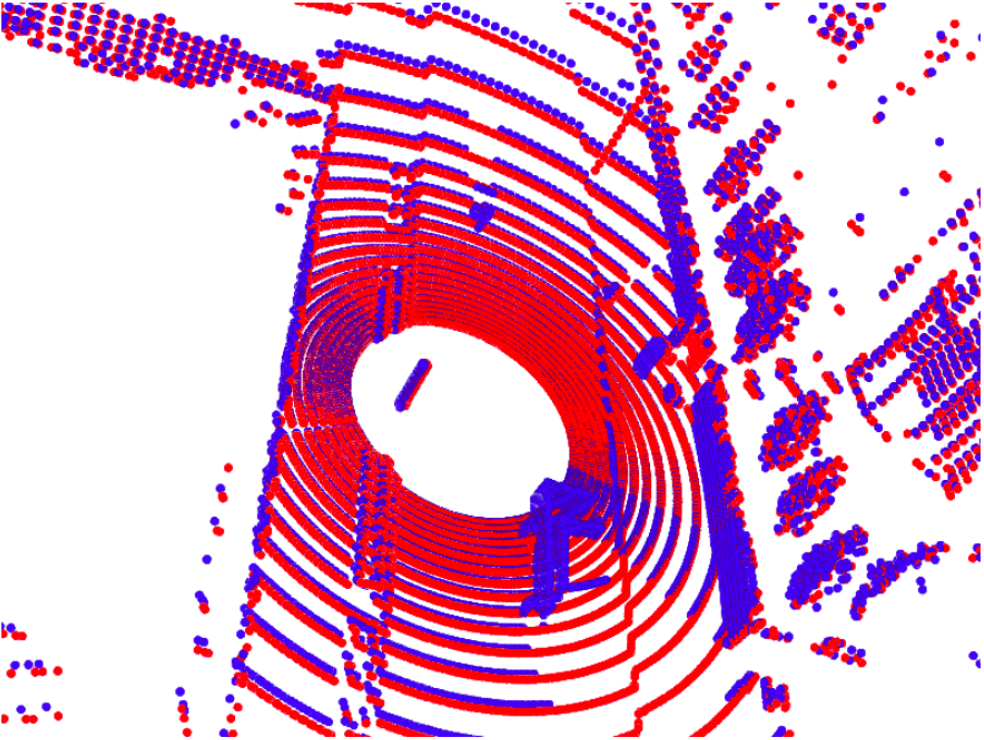}}
\end{subfigure}\hfil
\\[1ex]
\begin{subfigure}[t]{0.175\textwidth}
\centering
\fbox{\includegraphics[width=3cm,height=2cm]{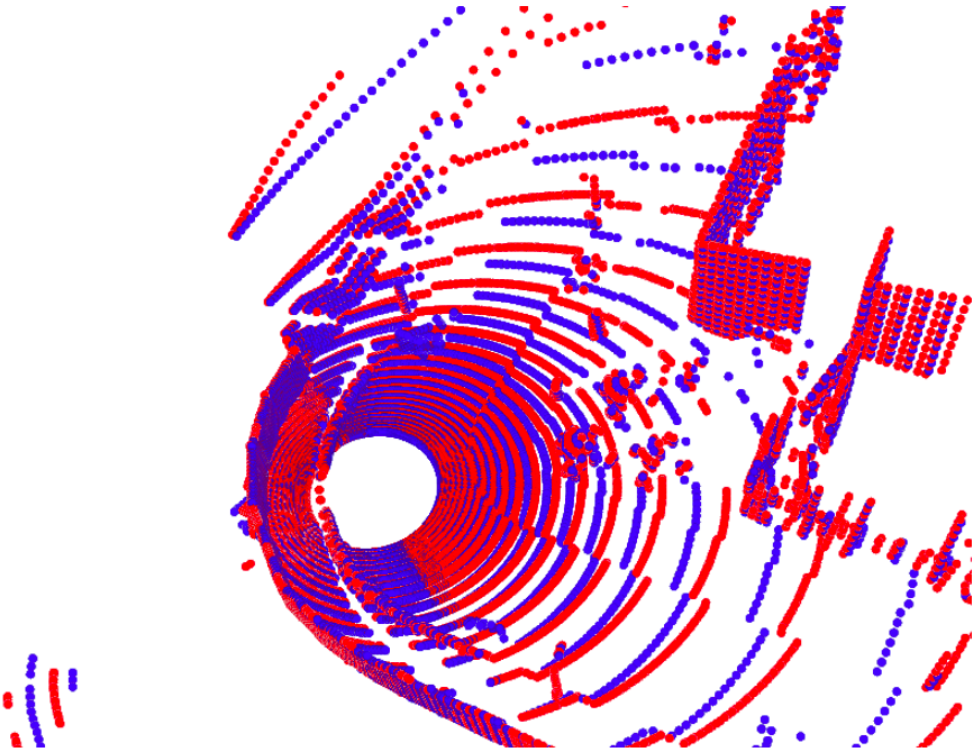}}
\end{subfigure}\hfil 
\begin{subfigure}[t]{0.175\textwidth}
\centering
\fbox{\includegraphics[width=3cm,height=2cm]{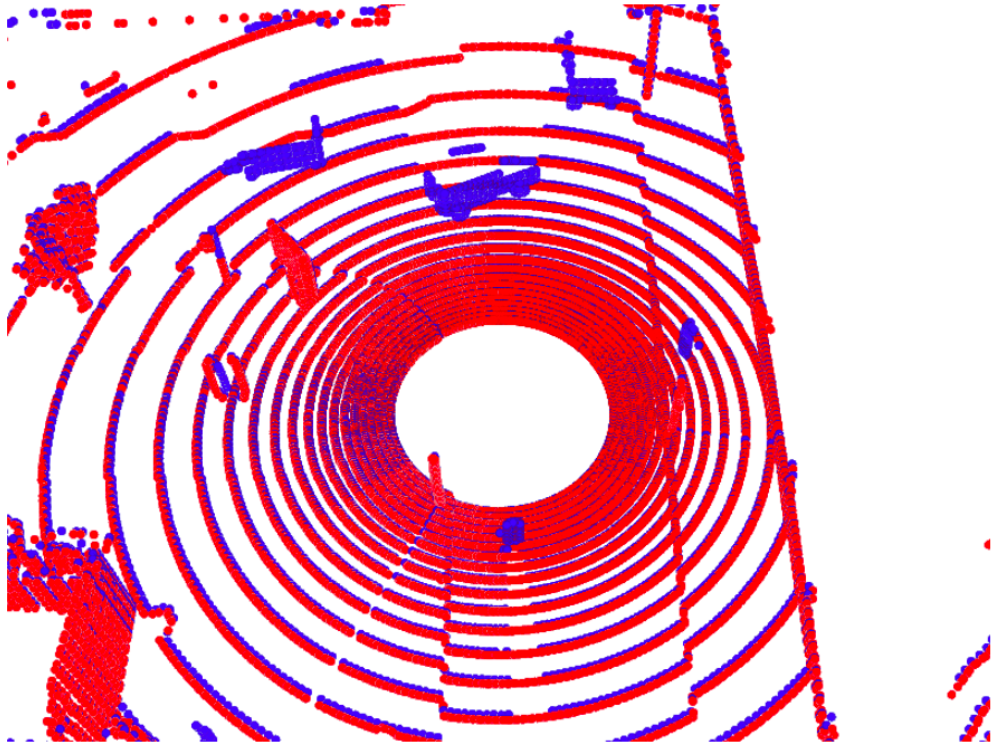}}
\end{subfigure}\hfil
\begin{subfigure}[t]{0.175\textwidth}
\centering
\fbox{\includegraphics[width=3cm,height=2cm]{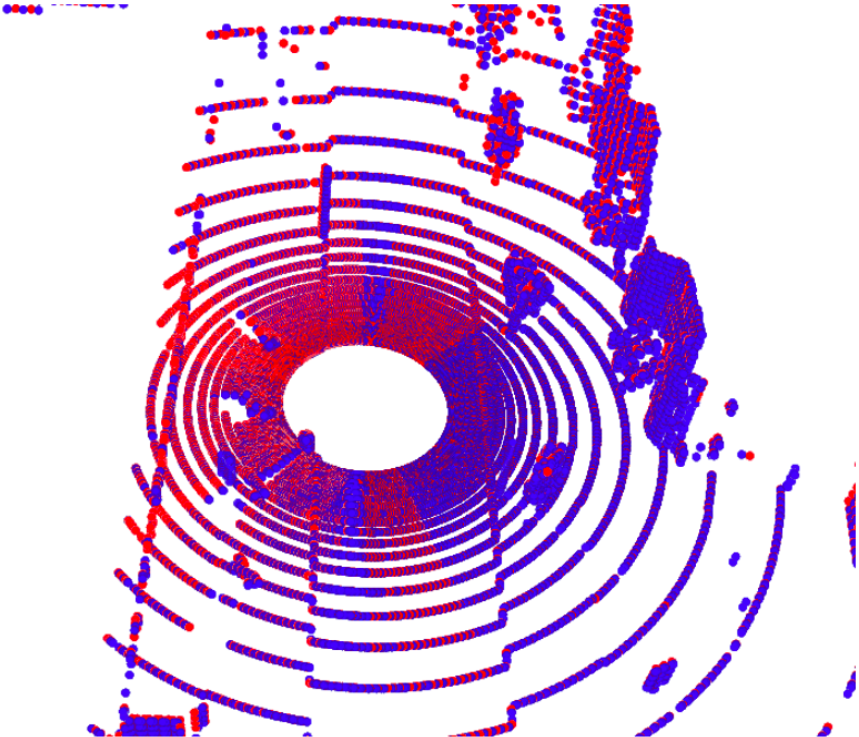}}
\end{subfigure}\hfil 
\begin{subfigure}[t]{0.175\textwidth}
\centering
 \fbox{\includegraphics[width=3cm,height=2cm]{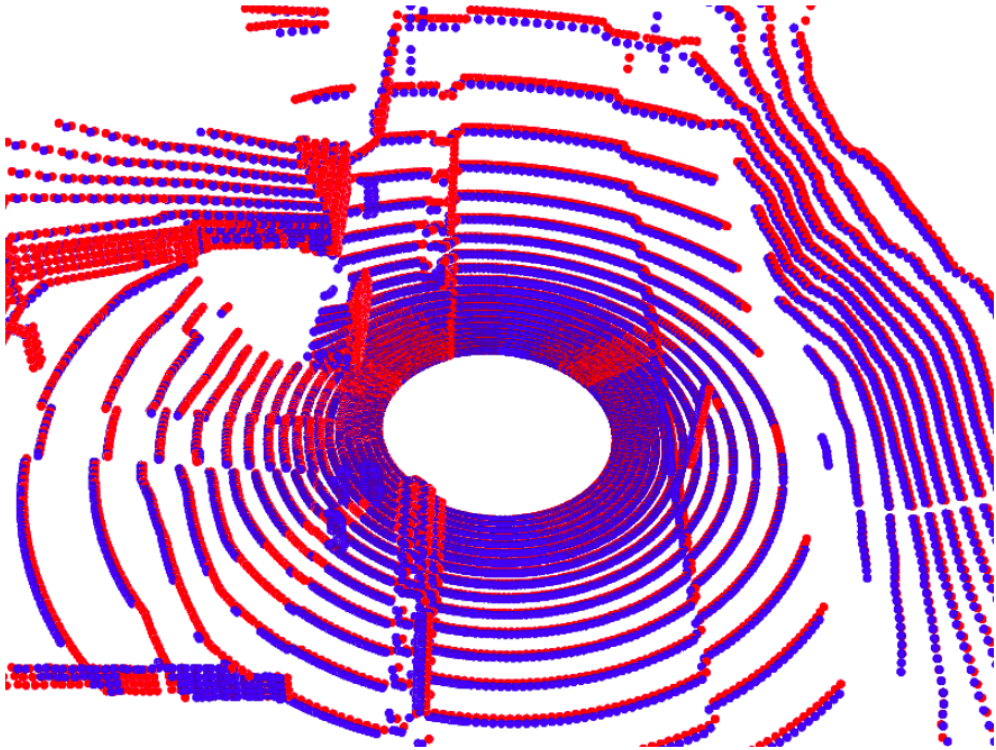}}
\end{subfigure}\hfil 
\begin{subfigure}[t]{0.175\textwidth}
\centering
 \fbox{\includegraphics[width=3cm,height=2cm]{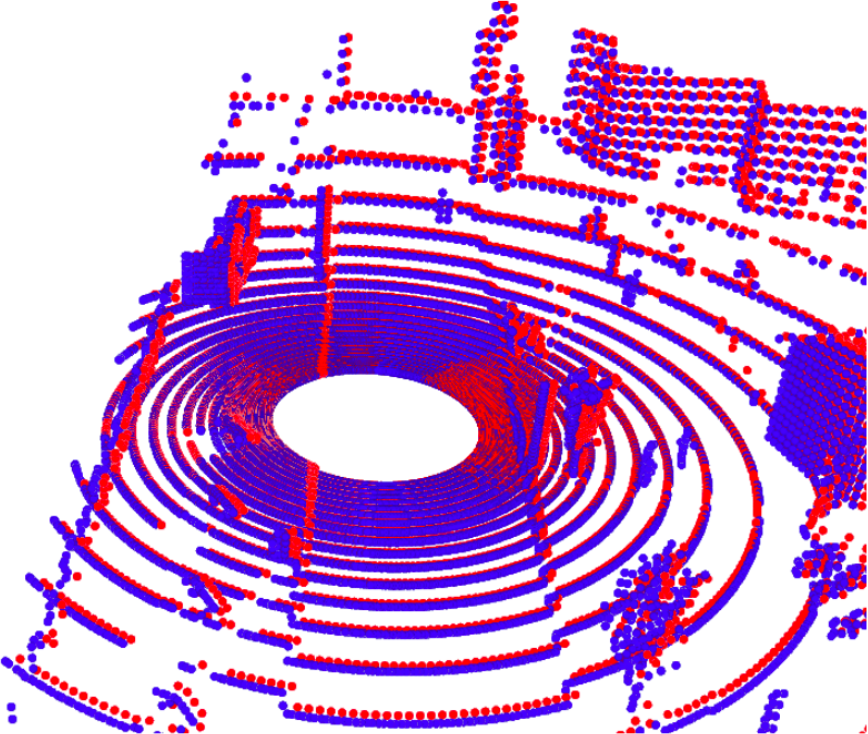}}
\end{subfigure}\hfil
 \captionsetup{belowskip=-15pt}
\caption{Examples of corresponding dynamic and static pairs from Carla-64 dataset. Best viewed in color. Dynamic and static scans are overlaid on each other to show the accuracy of correspondence. No segmentation information was used to create this dataset. Here, blue denotes dynamic scan and red denotes static scan. Regions with only red points indicate regions that are occluded in blue scans.}
\label{dataset_images}

\end{figure*}

\begin{figure}[t]
\subfloat[Without Segmentation]{
  \includegraphics[width=120pt]{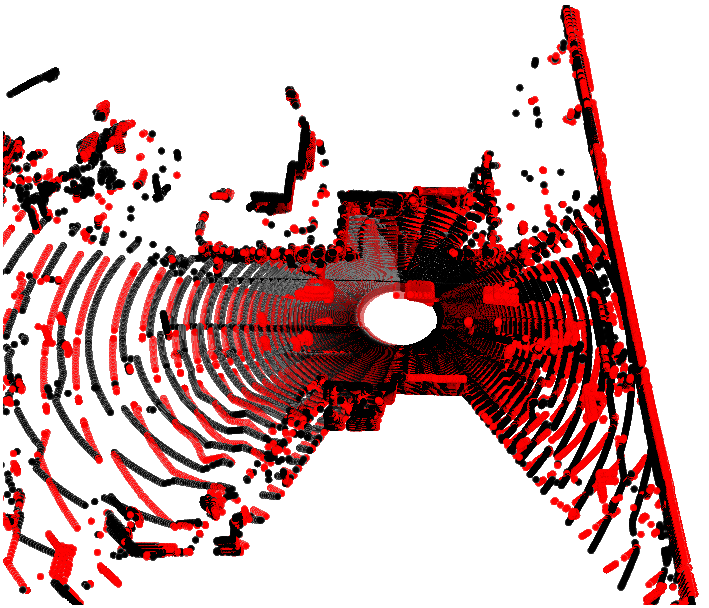}%
  \includegraphics[width=120pt]{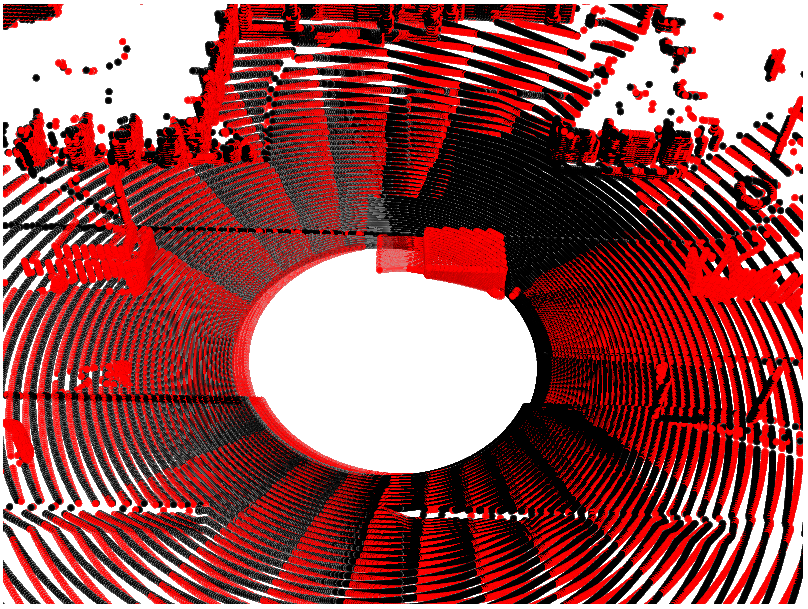}
}

\subfloat[With Segmentation]{%
  \includegraphics[width=120pt]{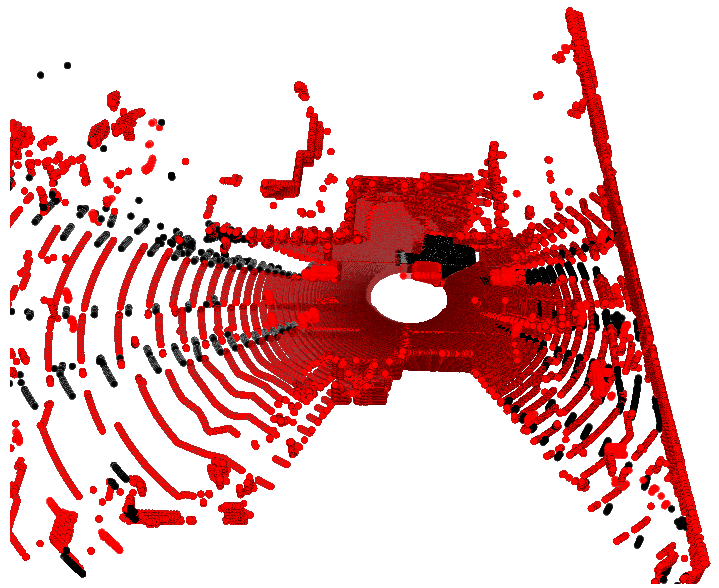}%
  \includegraphics[width=120pt]{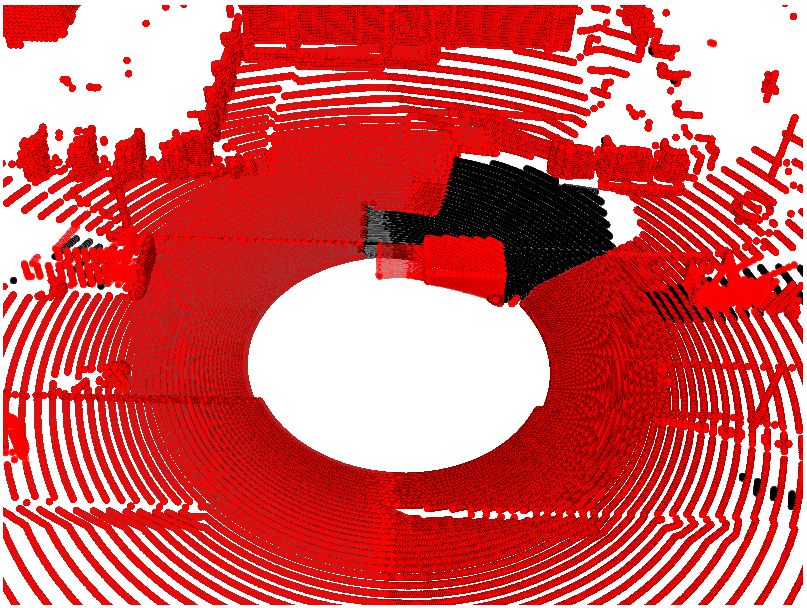}
}
\caption{Paired Correspondence Dataset created with and without segmentation. Best viewed in color. The right image is zoomed-in version of left image. Here, red denotes dynamic scan and black denotes corresponding static scan in the dataset. We can see that for dataset without segmentation, there is some mismatch between static points in both the scans. In the dataset using segmentation (which was used to train \DSLRpp{}), we can see no such mismatches and the static points in both the scans perfect overlay on each other.}
\label{fig:dslrplus_images}
\end{figure}

We list the details of adversarial training in the algorithm. We use a value of $\alpha$ = 10 for the dual loss in the training procedure. The training procedure has been described in the Algorithm \ref{AdversarialAlgo}.

\subsubsection{Inference}
\DSLR{} inference is real time. \DSLR{} takes ~11ms for inference on a GTX 1080Ti while Velodyne LiDAR (used in all our datasets) take ~100ms to generate a complete 360\textdegree scan.

\subsection{Dataset Generation}
\label{gen_data}
Standard Point clouds e.g. ShapeNet\cite{chang2015shapenet} are different in structure and complexity from LiDAR point clouds. The number of points in a single VLP-64 LiDAR frame running at a frequency of 10Hz is $\approx$ 1,30,000, whereas a standard point cloud in ShapeNet dataset has $\approx $ 2600 points \cite{chang2015shapenet}. LiDAR point clouds capture information at distances as far as 120m which is not the case with standard point clouds. Moreover, LiDAR point clouds have rich geometric and 3D information of a scene which is not the case with the arbitrary point clouds. Therefore we argue that using explicit static supervision for corresponding dynamic LiDAR frames is vital for a holistic reconstruction of a dynamic LiDAR scans to its corresponding static counterpart which may not be the case for standard arbitrary point clouds.\\
We detail an algorithm that extracts corresponding static and dynamic pairs of LiDAR scans using static and dynamic runs in the same environment using the same path. We collect such data from CARLA Simulator as well as using our slow-moving Unmanned Ground Vehicle (UGV). The dataset generation method has also been explained in Algorithm\ref{corr_dataset_algo}.

\subsubsection{Dataset Generation with segmentation for \DSLRpp{}}
Even after applying relative pose transformation in dataset generation algorithm, there can be some mismatch between static LiDAR scan points. This occurs because the dynamic and corresponding LiDAR scans were recorded at different places in the environment. Relative pose transformation can ensure that the static structures like walls overlap in both the scans but sometimes the exact LiDAR points don't overlap. Figure \ref{fig:dslrplus_images}shows this difference between the dynamic and corresponding static frame.

This can lead to poor model performance since the model now is forced to learn on somewhat noisy static points. If segmentation information is available, we can overcome this issue using a technique similar to \DSLRSeg{}. Note that the corresponding static and dynamic frames have a degree of similarity equivalent to the percentage of static points in the dynamic frame.  Table \ref{kitti_dynamism} shows that typically in real-world scenarios the amount of dynamism doesn't exceed  10\%. This knowledge is leveraged to construct more accurate training data.

We use segmentation information to generate a mask to identify the points falling on dynamic objects. Using this mask, we only take regions corresponding to dynamic objects from corresponding static and regions corresponding to static points from the ground truth dynamic. Figure \ref{fig:dslrplus_images}shows that corresponding static generated by this method has less error and all the static LiDAR points align perfectly in both the scans. 

\subsubsection{Ati Real-world Dataset (ARD)-16 generation}
 We create first of its kind real world dynamic-static paired correspondence dataset collected using VLP-16 LiDAR. It was captured in outdoor environment at Robert Bosch centre, IISc with no moving objects during static run and several moving objects (1 car, 1 2-wheeler, few pedestrians) during dynamic run. It consists of ~1.5k scans/run and we collected 10 dynamic and 5 static runs. This gave ~14k LiDAR scan pairs for train, val and test. For busy urban areas, static runs can be obtained at night time.

\begin{table*}[t]
\centering
\fontsize{9pt}{\baselineskip}\selectfont
\setlength\extrarowheight{2pt}
\caption{Comparison of the reconstruction baselines for various datasets. Lower is better.}
\label{table:mainresults}
\begin{tabular}{c|c|ccc|cc|c}
\hline
Model & Uses Seg & \multicolumn{3}{c}{Carla-64} & \multicolumn{2}{c}{ARD-16} & KITTI-64  \\\hline
 & & EMD & Chamfer & LQI & EMD & Chamfer & LQI\\
 \cline{3-8} 
AtlasNet  & No & 5681.85 & 5109.98 &- & 1464.61 & 176.46 & -\\
ADMG      & No & 397.94 & 6.23 & 7.049 & 309.64 & 1.62 & 2.911 \\
CHCP-VAE  & No & 343.98 & 9.58 & 4.080 & 88.94 & 0.67 & 1.128\\
CHCP-GAN & No & 329.38 & 8.19 & 3.519 & 65.24 & 0.38 & 1.133\\
CHCP-AE & No & 253.91 & 4.05 & 3.720 & 65.40 & 0.31 & 1.738\\
WCZC  & Yes & $2.73*10^{6}$ & 478.12 &- & - & - & - \\
EmptyCities  & Yes & 640.97 & 29.39 &- & - & - & - \\
\hline
DSLR (Ours) & No & \textbf{232.51} & \textbf{1.00} & \textbf{3.350} & \textbf{57.75} & \textbf{0.20} & 1.120\\
DSLR++ (Ours) & Yes & 205.48 & 0.49 & - & - & - \\
DSLR-Seg (Ours) & Yes & \textbf{150.90} & \textbf{0.02} & - & - & -\\
DSLR-UDA(Ours) & Yes & - & - & - & - & - & \textbf{1.119} \\
\hline
\end{tabular} 
\end{table*}

\subsection{LiDAR Reconstruction Evaluation Metrics}
We evaluate the difference between model reconstructed static and ground truth static LiDAR scan on CARLA-64 and ARD-16 datasets using the following two metrics:
\begin{itemize}

    \item \textbf{EMD}: Earth Mover's Distance is the minimum cost for transforming from one point cloud to the other and is given by:
    \begin{equation}
    \begin{split}
        dis_{\mathrm{EMD}}\left(P_{1}, P_{2}\right)=\min _{\eta: P_{1} \longrightarrow P_{2}} \sum_{x \in S_{1}}\|x-\eta(x)\|_{2}
        \end{split}
    \end{equation}
    Here, $\eta$ represents 1-1 mapping between $P_{1}$ and $P_{2}$.
    
    \item \textbf{CD}: Chamfer Distance tries to capture the average mismatch between points in two given point clouds and is given by:
    \begin{equation}
        \begin{split}
      dis_{\mathrm{CH}}=\sum_{x \in P_{1}} \min _{y \in P_{2}}\|x-y\|_{2}^{2}+\sum_{y \in P_{2}} \min _{x \in P_{1}}\|x-y\|_{2}^{2}
      \end{split}
    \end{equation}
    Here, $P_{1}$ and $P_{2}$ represents 2 sets of 3D points.

\end{itemize}
KITTI-64 does not have corresponding ground static, thus we propose a new metric LiDAR scan Quality Index LQI which estimates the quality of reconstructed static LiDAR scans by explicitly regressing the amount of noise in a given scan. This has been adapted from the CNN-IQA model \cite{kang2014convolutional} which aims to predict the quality of images as perceived by humans without access to a reference image.

\subsection{Unsupervised Domain Adaptation}
Let ${x}_{s} \in X_{s}$ represent data points from the source domain, ${x}_{t} \in X_{t}$ represent data points from the target domain, and $f(.)$ is the function used to map the data to a reproducing kernel Hilbert space (RKHS). The MMD is empirically approximated as follows:
\begin{equation}
\begin{split}
MMD(X_s,X_t)=\Bigg\|\frac{1}{|X_s|} \sum_{{x}_s\in X_s}f({x}_s)- \frac{1}{|X_t|}\sum_{{x}_t\in X_t} f({x}_t)\Bigg\|
\end{split}
\end{equation}

\subsection{LiDAR scan Quality Index (LQI)}
LQI has been inspired by the work done on No-Reference Image Quality Assessment where the visual quality of an image is evaluated without any reference image or knowledge about the type of distortion present in the image. We would like to adopt this task for LiDAR scans but evaluating LiDAR scans on their visual quality cannot a useful metric as it does not align with human visual perception nor it is the end goal for LiDAR reconstructions to be visually better. Therefore, in LQI, we evaluate reconstructed static LiDAR scans on the basis of the amount of noise injected by the model during the reconstruction. The assumption we use here is that this noise is Gaussian and hence, the LQI is a measure of the variance of the gaussian noise present in a normalized LiDAR scan. In this section, the use of the word noise would imply gaussian noise.

The model used for LQI is adopted from \cite{kang2014convolutional}. The input LiDAR scan is normalized and passed through a convolutional layer with stride:-1 and a 7x5 kernel. This layer produces 50 feature maps, this is followed by a max pool and a min pool operation, after which each feature map is reduced to one max value and one min value. Two fully connected layers of 800 nodes after the pooling operation lead to a 1-dimensional output which is the LQI metric:- a measure of the variance of the noise present in the scan. The lower the LQI the lesser the noise. For training, Gaussian noise with varying parameters was added to LiDAR scans on the fly, and the model was expected to regress this variance and trained using the L1 loss between the predicted variance and the actual variance. Separate LQI models were trained using LiDAR scans from the KITTI-64 dataset and CARLA-64 dataset to score reconstructions on each dataset respectively.

The LQI can be considered as a good proxy to measure the quality of reconstruction for LiDAR scans for cases where ground truth is not available as we have already shown in the main paper that it correlates positively with Chamfer distance.

\begin{figure}[t]
\subfloat[Raw dynamic scan]{
  \includegraphics[width=120pt]{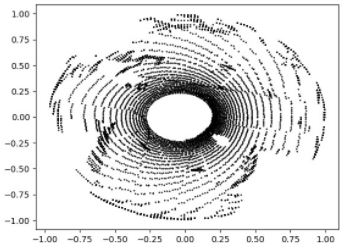}%
  \includegraphics[width=120pt]{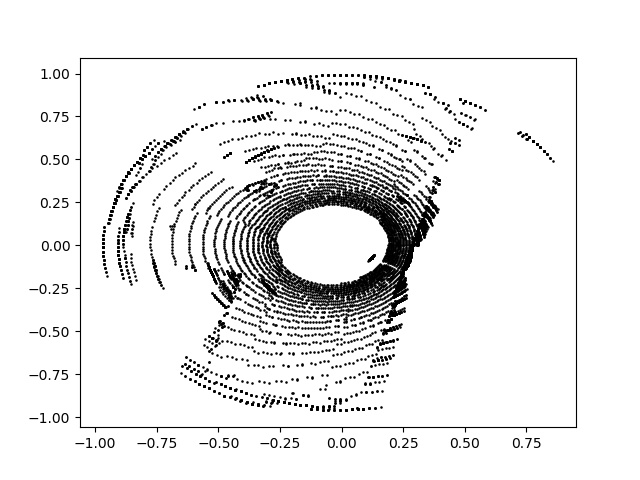}
}

\subfloat[Reconstructed Static Scan]{%
  \includegraphics[width=120pt]{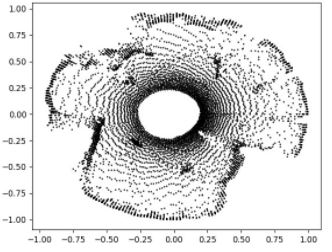}%
  \includegraphics[width=120pt]{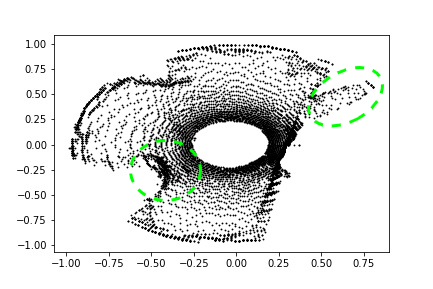}
}
\caption{Failure Case at Turns. The reconstruction obtained using \DSLR{} is not able to reconstruct the details in the dynamic scan. Edges are thick instead of sharp detail as indicated in the encircled regions, as well as the general clarity and crispness is missing.}
\label{fig:turnfail}
\end{figure}

\subsection{System Configuration and Training}
For our experiments we use an Intel Core i9-9900K CPU @ 3.60GHz processor,  loaded with NVIDIA GTX-2080Ti GPU. The autoencoder was trained for 600 epochs, the discriminator for 20 epochs while the adversarial framework was trained for 50 epochs.  We use Adam Optimizer with a initial learning rate of \emph{$6e-4$}, momentum of \emph{0.1}, and weight decay of \emph{$1e-5$}. 

The LQI model was trained for 9 epochs. Adam optimizer was used with the initial learning rate of \emph{$1e-2$}, momentum of \emph{0.1} with learning rate reducing by 0.5 after every 3rd epoch.

\section{More LiDAR Reconstruction Results}
\subsection{Failure cases}
We observe that reconstruction performs well when lateral movement of the vehicle is limited. Moderately occluded scenes are also reconstructed clearly in such cases. However, for cases like turns, road intersections model does not give a very accurate reconstruction. We also observe that it's hard for such a reconstruction algorithm to be generalizable. Testing with scans of a totally new region degrades the static reconstruction. This is because LiDAR scans in one region or area might be completely different from other areas and given the model has seen LiDAR scans of only a particular region or area, it might give sub-optimal results on input of a totally different  region. Fig. \ref{fig:turnfail} depicts the reconstruction at a turn. It is noticeable that LiDAR scan at turns are very different from normal straight motion scans and we observe that picking detailed structures is an issue in such cases.

\section{SLAM Results and Analysis}

Autonomous navigation systems which heavily rely on SLAM algorithms to localize and map in a known or unknown environment. Typically, SLAM algorithms are designed with the assumption that the environment is static. Violation  of this assumption in real world like the presence of dynamic obstacles leads to a poor SLAM performance. 

\begin{table}[b!]
    
    \caption{SLAM error (ATE in in m) in static and dynamic environments}
        \label{table:slamerr}
        \scalebox{0.9}{
        \begin{tabular}{llll}
            \hline
                    & w/o Loop & w/ Loop & w/ Loop closure \\
                    & closure  & closure & \& tuned \\
            \hline
            Dynamic & 44.4          & 30.6          & 6.15                \\
            Static  & 13.3 & 11.4 & 4.07       \\
            \hline
        \end{tabular}}\\
\end{table}

\begin{figure*}[t!]

    \centering
    \fbox{\includegraphics[scale=0.25]{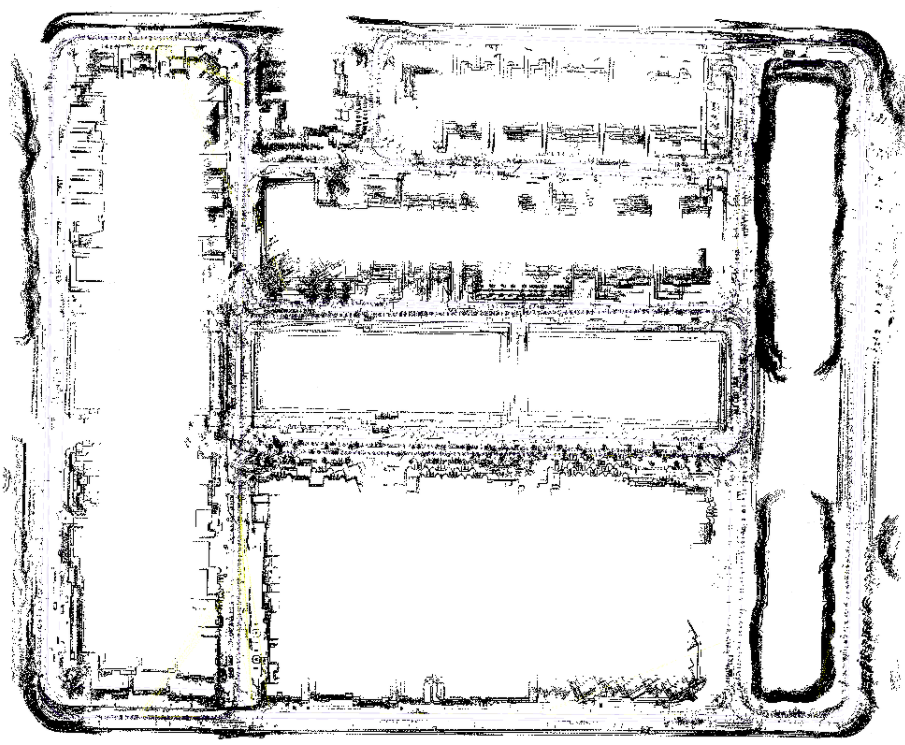}}
    \fbox{\includegraphics[scale=0.245]{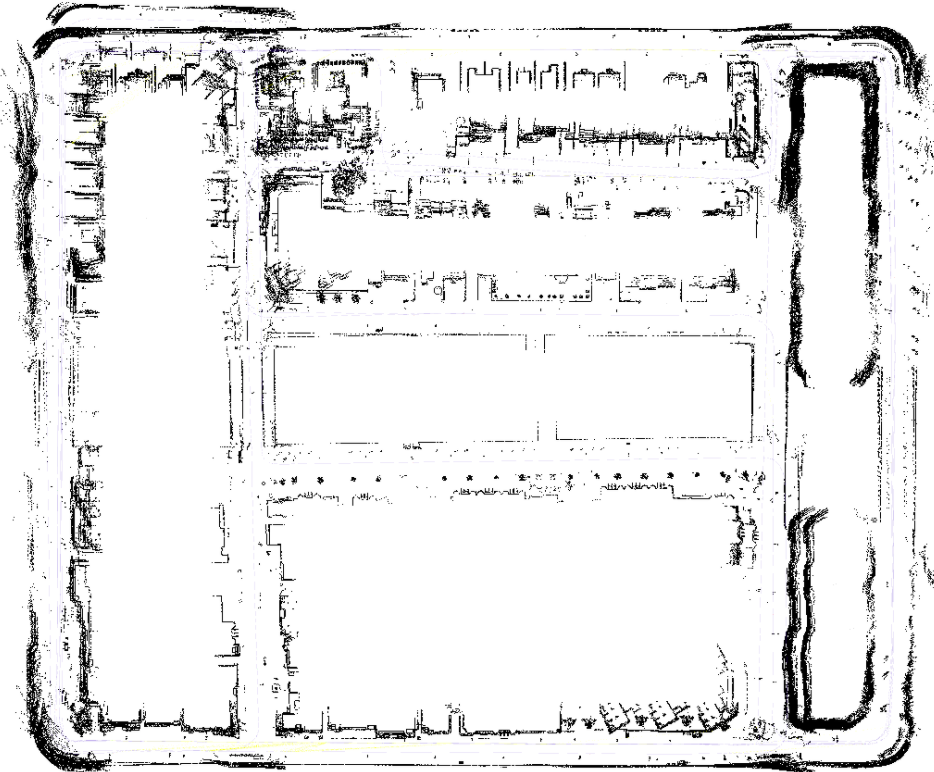}}
     \captionsetup{belowskip=-10pt}
    \caption{Map generated by SLAM algorithm (Cartographer) in the same CARLA environment for (1) Left: dynamic run (2) Right: static run. Although both the generated maps are not perfect, we can see that map generated from dynamic run has more corruptions compared to the map created from static run. Heavy corruptions in the map of an environment can result in degradation of localization performance over the time.}
    \label{dynamic_static_map}
\end{figure*}

We confirm that dynamic environment results not only in high SLAM error as also seen in Table \ref{table:slamerr} but also creates maps of the environment with corruptions due to dynamic objects as shown in figure \ref{dynamic_static_map}. The results for Table \ref{table:slamerr} were generated by running graph SLAM based LiDAR SLAM algorithm, Cartographer \cite{hess2016real} on the dynamic and static runs in the same environment of our CARLA-64 dataset. Loop closure in graph SLAM methods typically helps in reducing SLAM error (like Absolute Trajectory Error (ATE)) in dynamic environments but we show in Table \ref{table:slamerr} that even with loop closure in Cartographer, SLAM error in dynamic environment is two fold compared to static environment. Manually tuning the parameters in Cartographer helps reduce the problem but doesn't solve it completely.

\begin{table}[b!]
\centering
\caption{Detailed Comparison of DSLR-Seg with existing baselines on CARLA-64.} 
\label{table:carla}
\begin{adjustbox}{width=\columnwidth,center}
\begin{tabular}{cllllll}
\hline\noalign{\smallskip}
Carla Run & Model & ATE & drift & RPE Trans & RPE Rot \\
\hline
\multirow{5}{*}{1} & Pure-Dynamic  & 10.57 & 25.91 & 0.06 & 0.43\textdegree \\ 
& Detect \& Delete (Model-Seg)  & 24.15 & 44.47 & 0.06 & 0.43\textdegree  \\ 
& Detect\& Delete (GT-Seg) & 25.64 & 46.60 & 0.09 & \textbf{0.42\textdegree}  \\ 
& DSLR-Seg (Model-Seg) (Ours) & 23.04 & 41.7 &\textbf{0.057} & 0.43\textdegree \\
& DSLR-Seg (GT-Seg) (Ours) & \textbf{8.53} & \textbf{22.08} &0.06 & 0.42\textdegree \\

\hline
\multirow{5}{*}{2} & Pure-Dynamic  & 4.72 & 4.25 & 0.05 & 0.41\textdegree \\ 
& Detect \& Delete (Model-Seg) & 2.49 &  2.61 & 0.062 & 0.50\textdegree  \\ 
& Detect \& Delete (GT-Seg) & 2.49 &  \textbf{2.50} & 0.058 & 0.41\textdegree  \\ 
& DSLR-Seg (Model-Seg) (Ours) & 2.14 & 2.8 & 0.05 & 0.38 \textdegree \\
& DSLR-Seg (GT-Seg) (Ours) & \textbf{2.07} & 2.68 & \textbf{0.05} & \textbf{0.24 \textdegree} \\

\hline
\multirow{5}{*}{3} & Pure-Dynamic  & 15.79 & 25.58 & 0.06 & 0.37\textdegree \\ 
& Detect \& Delete (Model-Seg) & \textbf{4.66} & 9.53 & 0.06 & 0.37\textdegree \\ 
& Detect \& Delete (GT-Seg) & 6.84 & 21.20 & 0.05 & 0.37\textdegree \\ 
& DSLR-Seg (Model-Seg) (Ours) & 7.12 & \textbf{7.8} & 0.05 & 0.37\textdegree\\
& DSLR-Seg (GT-Seg) (Ours) & 11.39 & 16.15 & \textbf{0.05} & \textbf{0.37}\textdegree\\ 
\hline
\end{tabular}
\end{adjustbox}
\end{table}

\begin{figure}[t!]
    \centering
    \includegraphics[scale=0.4]{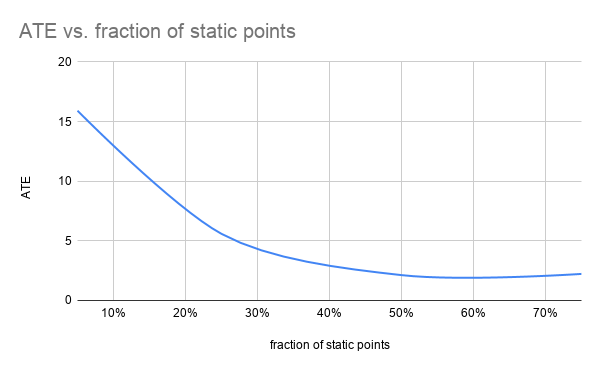}
    \caption{We experimentally study here the effect of reduced percentage of static points in a LiDAR scan. The fraction of static points were varied on X axis by randomly deleting points of individual LiDAR scans of a static run from CARLA-64 dataset. For a given fraction of static points, we evaluate SLAM error or Absolute Trajectory Error (ATE). It is seen that SLAM error doubles as the percentage of static points from a scene falls below 50\%.}
    \label{fig:static_point_validation}
\end{figure}

 Therefore, instead of deleting these dynamic points, we propose to reconstruct the dynamic occlusion holes with the background static points which can significantly improve SLAM performance as per the figure \ref{fig:static_point_validation}. The metrics used for evaluation are described below

\begin{itemize}
    \setlength\itemsep{0em}
    \item \textbf{ATE} or Absolute Trajectory Error measures the similarity between shape of pose trajectory estimated by SLAM algorithm and is computes similar to \cite{6385773}.
    \item \textbf{RPE} or Relative Pose Error also measures the pose error between two trajectories estimated by SLAM algorithm but it also is robust to accumulated global error. It is computed similar to \cite{6385773}. \textit{RPE} is computed differently for translational and rotational errors and is referred to here as \textbf{RPE trans} and \textbf{RPE rot}. 
    \item \textbf{Drift} is the average of individual pose error w.r.t. ground truth pose and measures how far the robot is from its actual ground truth pose. Unattended high drift in robots can often lead to collision with surrounding objects.
\end{itemize}

We compare our proposed joint model approach for SLAM with existing baselines, which detect the dynamic points and delete it \cite{ruchti2018mapping}, \cite{vaquero2019improving}, \cite{li2019net},\cite{nagy2018real} which we term as \textbf{Detect \& Delete} in our results and with the base case with no LiDAR preprocessing namely \textbf{Dynamic}. 

Our \DSLRSeg{} model and Detect \& Delete baseline use dynamic segmentation to detect dynamic points, we report results with two variations of segmentations. They are reported with Model-Seg variation if U-net was used to give segmentation masks for dynamic points or GT-Seg variation is ground truth available mask was used segmentation. While Model-Seg variation shows the effectiveness of baseline and our method on a live vehicle, GT-Seg shows the effectiveness of Detect \& Delete and our \DSLRSeg{} model in an ideal segmentation scenario where all the dynamic points of a given frame are perfectly known.

\begin{table}[t!]
\centering
\caption{Static dynamic points in each KITTI-64 run}
\label{table:static_dynamic}
\begin{tabular}{clllllllll}
\hline\noalign{\smallskip}
 Run &  S\% & SD\% & MD\% \\
\noalign{\smallskip}
\hline
\noalign{\smallskip}
\multirow{1}{*}{0}   & \multirow{1}{*}{91.07\%} & \multirow{1}{*}{8.82\%} & \multirow{1}{*}{0.09\%}\\
\multirow{1}{*}{1} &   \multirow{1}{*}{99.15\%} & \multirow{1}{*}{0\%} & \multirow{1}{*}{0.8\%}\\
\multirow{1}{*}{2} &   \multirow{1}{*}{97.6\%} & \multirow{1}{*}{2.3\%} & \multirow{1}{*}{0.07\%}\\
\multirow{1}{*}{4}  & \multirow{1}{*}{98.69\%} & \multirow{1}{*}{0.36\%} & \multirow{1}{*}{0.93\%}\\
\multirow{1}{*}{5} &  \multirow{1}{*}{96.51\%} & \multirow{1}{*}{3.18\%} & \multirow{1}{*}{0.29\%}\\
\multirow{1}{*}{6} &   \multirow{1}{*}{93.22\%} & \multirow{1}{*}{6.68\%} & \multirow{1}{*}{0.09\%}\\
\multirow{1}{*}{7} &  \multirow{1}{*}{89.2\%} & \multirow{1}{*}{9.96\%} & \multirow{1}{*}{0.08\%}\\
\multirow{1}{*}{9} & \multirow{1}{*}{95.55\%} & \multirow{1}{*}{4.21\%} & \multirow{1}{*}{0.23\%}\\
\multirow{1}{*}{10} & \multirow{1}{*}{98.03\%} & \multirow{1}{*}{1.6\%} & \multirow{1}{*}{0.36\%}\\
\hline
\end{tabular}
\label{kitti_dynamism}
\end{table}

\begin{table}[t!]

\caption{Detailed Comparison of \DSLRSeg{} with existing Baselines on KITTI-64.}
\label{table:kitti}
\centering
\begin{adjustbox}{width=\columnwidth,center}
\begin{tabular}{clllll}
\hline\noalign{\smallskip}
 Run &  Model & ATE & drift & RPE trans & RPE rot \\
\noalign{\smallskip}
\hline
\noalign{\smallskip}

\multirow{5}{*}{0} & Pure-Dynamic  &  \textbf{7.137} & 6.420 & 1.217 & \textbf{1.772}\textdegree \\ 

&  Detect \& Delete (Model-Seg) & 7.368 & 5.345 & 1.217 & 1.774\textdegree  \\ 
&  Detect \& Delete (GT-Seg) & 12.088 & 17.913 & 1.217 & 1.776\textdegree  \\ 
&   DSLR-Seg (Model-Seg) (Ours) & 7.250 & \textbf{4.530} & 1.217 &  1.771\textdegree \\
&  DSLR-Seg (GT-Seg) (Ours) & 7.831 & 5.594 & \textbf{1.217} &  1.776\textdegree \\
\hline

\multirow{5}{*}{1} & Pure-Dynamic  & 40.229 & 68.429 & 3.187 & 1.155\textdegree \\ 
&  Detect \& Delete (Model-Seg) & 39.627 & \textbf{54.891}   & 3.188 & 1.155\textdegree  \\ 
&   Detect \& Delete (GT-Seg) & 52.589 & 82.683 & 3.179 & 1.158\textdegree  \\ 
&   DSLR-Seg (Model-Seg) (Ours) & 146.648 & 169.260 & \textbf{3.088} &  1.157\textdegree \\ 
&   DSLR-Seg (GT-Seg) (Ours) & \textbf{33.232} & 60.215 & 3.204 &  \textbf{1.155}\textdegree \\
\hline

\multirow{5}{*}{2} & Pure-Dynamic &  37.915 & \textbf{28.623} & 1.555 & 1.430\textdegree \\ 
&   Detect \& Delete (Model-Seg) & 49.199 & 39.124 & 1.552 & 1.430\textdegree  \\ 
& Detect \& Delete (GT-Seg) & \textbf{12.483} & 69.892 & 1.557 & \textbf{1.429}\textdegree  \\ 
&  DSLR-Seg (Model-Seg) (Ours) & 75.948 & 110.594 & \textbf{1.547} &  1.430\textdegree \\ 
&  DSLR-Seg (GT-Seg) (Ours) & 47.012 & 87.193 & 1.558 &  1.430\textdegree \\
\hline

\multirow{5}{*}{4}  & Pure-Dynamic  &  0.565  & \textbf{0.715} & 2.055 & 0.192\textdegree \\ 
&  Detect \& Delete (Model-Seg) & 0.268 & 1.078 & 2.050 & 0.192\textdegree  \\ 
&   Detect \& Delete (GT-Seg) & 0.588 & 1.026 & 2.049 & 0.192\textdegree  \\ 
&   DSLR-Seg (Model-Seg) (Ours) & 0.598 & 1.208 & 2.054 &  0.192\textdegree \\
&   DSLR-Seg (GT-Seg) (Ours) & \textbf{0.218} & 0.785 & \textbf{2.049} &  \textbf{0.192}\textdegree \\
\hline

\multirow{5}{*}{5} & Pure-Dynamic  & 8.798 & 11.103 & 1.209 & 1.261\textdegree \\ 
&  Detect \& Delete (Model-Seg) & 2.507 & 3.335 & 1.207 & 1.262\textdegree  \\ 
&  Detect \& Delete (GT-Seg) & \textbf{1.815} & 2.276 & 1.209 & 1.264\textdegree  \\ 
& DSLR-Seg (Model-Seg) (Ours) & 5.946 & 6.725 & \textbf{1.204} &  \textbf{1.261}\textdegree \\
&  DSLR-Seg (GT-Seg) (Ours) & 2.217 & \textbf{2.211} & 1.208 &  1.264\textdegree \\
\hline

\multirow{5}{*}{6} & Pure-Dynamic  & 1.783 & 2.576 & 1.632 & 1.514\textdegree \\ 
&   Detect \& Delete (Model-Seg) & 2.941 & 4.891 & 1.631 & 1.514\textdegree  \\ 
&  Detect \& Delete (GT-Seg) & \textbf{1.748} & \textbf{1.631} & \textbf{1.515} & 1.776\textdegree  \\ 
&   DSLR-Seg (Model-Seg) (Ours) & 2.424 & 4.281 & 1.632 &  1.514\textdegree \\
&  DSLR-Seg (GT-Seg) (Ours) & 3.687 & 4.148 & 1.630 &  \textbf{1.514}\textdegree \\
\hline

\multirow{5}{*}{7} & Pure-Dynamic  & 1.429 & 2.777 & \textbf{1.025} & \textbf{1.646}\textdegree \\ 
&  Detect \& Delete (Model-Seg) &  1.205 & 1.982 & 1.025 & 1.665\textdegree \\ 
&  Detect \& Delete (GT-Seg) & 12.088 & 17.913 & 1.217 & 1.776\textdegree  \\ 
&  DSLR-Seg (Model-Seg) (Ours) & \textbf{1.075} & \textbf{1.555} & 1.028 & 1.665\textdegree \\ 
&  DSLR-Seg (GT-Seg) (Ours) & 1.537 & 2.798 & 1.026 &  1.663\textdegree \\
\hline

\multirow{5}{*}{9} & Pure-Dynamic  & \textbf{6.563} & 8.628 & 1.551 & 1.352\textdegree \\ 
&  Detect \& Delete(Model-Seg) & 7.732 & 7.544& 1.551 & 1.353\textdegree \\
&  Detect \& Delete(GT-Seg) & 7.995 & 9.436 & \textbf{1.550} & \textbf{1.352}\textdegree  \\ 
&  DSLR-Seg (Model-Seg) (Ours) & 7.136 & \textbf{5.549} & 1.551 & 1.352\textdegree \\ 
& DSLR-Seg (GT-Seg) (Ours) & 7.647 & 7.656 & 1.552 &  1.353\textdegree \\
\hline

\multirow{5}{*}{10} & Pure-Dynamic & 1.866 & 5.498 & 1.195 & 1.308\textdegree \\ 
&  Detect \& Delete (Model-Seg) &  \textbf{1.629} & 4.208 & \textbf{1.192} & \textbf{1.303}\textdegree \\ 
& Detect \& Delete (GT-Seg) & 1.732 & \textbf{2.938} & 1.194 & 1.305\textdegree  \\ 
&  DSLR-Seg (Model-Seg) (Ours) & 2.409 & 5.652 & 1.194 & 1.307\textdegree \\ 
&  DSLR-Seg (GT-Seg) (Ours) & 1.826 & 6.501 & 1.196 &  1.308\textdegree \\
\hline
\end{tabular}
\end{adjustbox}
\end{table}

Table \ref{table:carla} reports the results of our method \DSLRSeg{} on Carla-64 dataset. The \DSLRSeg{} model was trained on this dataset and is finally evaluated on 3 long runs. We primarily test on runs containing loops because we would like our methods to use loop closure to correct for drifts developed over time. We wanted to show that SLAM algorithms fail in dynamic environments even with loop closure detection and our proposed method can give significant improvements over these methods. Our algorithm is improving the quality of reconstructed scan in dynamic setting which ensures a better chance of loop closure due to increment in matching score used for place recognition in loop closure detection. We can see that \DSLRSeg{} always outperforms the base case Dynamic method and is mostly better than the baseline method in most of the metrics, especially in ATE and drift.

\setlength{\tabcolsep}{4pt}
\begin{table}[b!]
\centering
\caption{Detailed Comparison of DSLR on ARD-16 dataset.}
\begin{adjustbox}{width=\columnwidth,center}
\label{table:ati}
\begin{tabular}{clllll}
\hline\noalign{\smallskip}
Run & Model & ATE & RPE Trans & RPE Rot \\
\hline
\multirow{2}{*}{1} & Pure-Dynamic  & 2.142 & 0.048 & 0.927\textdegree \\
& DSLR (Ours) & \textbf{2.117} & \textbf{0.045} & \textbf{0.904}\textdegree\\ 
\hline
\multirow{2}{*}{2} & Pure-Dynamic  & 6.939 & 0.051 & 0.823\textdegree \\
& DSLR (Ours) & \textbf{6.916} & \textbf{0.048} & \textbf{0.806}\textdegree\\ 
\hline
\multirow{2}{*}{3} & Pure-Dynamic  & 0.398 & 0.036 & \textbf{0.457}\textdegree \\
& DSLR (Ours) & \textbf{0.395} & \textbf{0.035} & 0.46\textdegree\\ 
\hline
\multirow{2}{*}{4} & Pure-Dynamic  & 0.230 & 0.027 & \textbf{0.613}\textdegree \\
& DSLR (Ours) & \textbf{0.220} & \textbf{0.028} & 0.612\textdegree\\ 
\hline

\multirow{2}{*}{5} & Pure-Dynamic  & 0.254 & 0.028 & \textbf{0.442}\textdegree \\
& DSLR (Ours) & \textbf{0.241} & \textbf{0.027} & 0.447\textdegree\\ 
\hline

\multirow{2}{*}{6} & Pure-Dynamic & 0.245 & 0.028 & \textbf{0.418}\textdegree \\
& DSLR (Ours) & \textbf{0.244} & \textbf{0.027} & 0.460\textdegree\\ 
\hline

\end{tabular}
\end{adjustbox}
\end{table}

We trained and evaluate our model \DSLR{} on ARD-16 dataset and report our results in table \ref{table:ati} against base case Dynamic. Since ARD-16 dataset was collected by us, we do not have ground truth segmentation mask. Therefore, we do not report SLAM performance for Detect \& Delete and \DSLRSeg{} that require dynamic segmentation. In these results, we would like to bring out the fact that how in real world setting, segmentation masks may not be readily available. Despite this, we can not only collect data and train our proposed model \DSLR{}, but it clearly outperforms the base case Dynamic in almost all the metrics especially ATE. However, we do note that the difference in result between \DSLR{} and Dynamic is not quite high. This is because all the LiDAR frames in this run repeatedly observe the same part of the environment and Cartographer in both the cases is able to quickly correct its SLAM errors using loop closures. However, in a city like environment (similar to CARLA-64 or KITTI-64 datasets) such all-to-all loop closing constraints are not found often and our model will perform better than base case Dynamic.

In table \ref{table:kitti}, we compare our method against Dynamic and baseline method. A key thing to note is that our model was not trained on this dataset and we are reporting results using the \DSLRSeg{} model trained on CARLA-64 dataset. Despite this, we can see that our model \DSLRSeg{} is able to generalize to unseen dataset like KITTI-64 and our \DSLRSeg{} framework performs comparable or better to baseline methods in most of the cases.

We also report the percentage of dynamism in each these runs in Table \ref{table:static_dynamic}. \textbf{S\%}, \textbf{SD\%}, \textbf{MD\%} refers to average percentage of static points, average percentage of stationary dynamic points and average percentage of moving dynamic points in a given run, respectively. One surprising result we see that there are few runs like run no. 0, 9 where the base case Dynamic performs better than the baseline and our method. This was also reported by \cite{chen2019suma++} and we can see that this happens when dynamic points are actually stationary during LiDAR frame capture. As a result, these dynamic runs get to see more number of static points compared to other methods and based on learnings from figure \ref{fig:static_point_validation} this helps Dynamic to be the best performing method.

We report another surprising result that in these kind of runs, like run number 6, SLAM error is minimum for Model-Seg variation of \DSLRSeg{} compared to GT-Seg variation. We can understand this because in these runs, any error in dynamic segmentation can help the Model-Seg variation see more number of static points compared to its GT-Seg variation counterpart. But we would still like to detect all kind of dynamic points and replace them with static background because replacing any kind of movable points from the map helping in long term SLAM \cite{vaquero2019improving,chen2019suma++}.



\bibliography{egbib}